\documentclass[lettersize,journal]{IEEEtran}
\usepackage{amsmath,amsfonts,amssymb}
\usepackage{array}
\usepackage[caption=false]{subfig}
\usepackage{textcomp}
\usepackage{stfloats}
\usepackage{url}
\usepackage{verbatim}
\usepackage{graphicx}
\usepackage{cite}
\usepackage{bbding}
\usepackage{pifont}
\usepackage{xcolor}
\usepackage{multirow}
\usepackage{booktabs}
\usepackage{tabularx}
\usepackage{hyperref}
\usepackage{algorithm}
\usepackage{algpseudocode}
\usepackage{parskip}
\usepackage{arydshln}
\usepackage{orcidlink}

\renewcommand\IEEEkeywordsname{Keywords}

\renewcommand{\figurename}{Figure}

\begin{document}
\title{AssistDLO: Assistive Teleoperation \\ for Deformable Linear Object Manipulation}

\author{
Berk~Güler$^{*}$~\orcidlink{0000-0002-7273-2441},
Simon~Manschitz~\orcidlink{0000-0001-6419-8157},
Kay~Pompetzki~\orcidlink{0000-0002-8448-4510},
Jan~Peters~\orcidlink{0000-0002-5266-8091}

\vspace{1em}

% Affiliations (using initials instead of superscript numbers)
\normalsize
B.~Güler, K.~Pompetzki, J.~Peters \\
\textit{TU Darmstadt, Institute for Intelligent Autonomous Systems, Computer Science Department, 64289 Darmstadt, Germany} \\
$^{*}$E-mail: berk@robot-learning.de \\

\vspace{0.5em}
B.~Güler, S.~Manschitz \\
\textit{Honda Research Institute Europe GmbH, Carl-Legien-Straße 30, 63073 Offenbach/Main, Germany} \\

\vspace{0.5em}
J.~Peters \\
\textit{German Research Center for AI (DFKI), Research Department: Systems AI for Robot Learning; Hessian.AI; Robotics Institute Germany (RIG); Centre for Cognitive Science, Hochschulstr. 10, 64293 Darmstadt, Germany}
}
\maketitle

\begin{abstract}
Manipulating Deformable Linear Objects (DLOs) is challenging in robotics due to their infinite-dimensional configuration space and complex nonlinear dynamics. 
In teleoperation, depth uncertainty hinders state perception and reaction. \textit{AssistDLO} addresses this challenge as an assistive teleoperation framework for DLO manipulation that combines real-time multi-view state estimation, visual assistance~(VA), and a geometry-aware shared-autonomy controller based on Control Barrier Functions~(SA-CBF). 
While traditional shared autonomy methods often rely on simple geometric attractors and may fail to preserve DLO geometry, SA-CBF acts as a geometry-aware funnel, facilitating precise grasping while preserving the operator's high-level authority. 
The framework is evaluated in a bimanual knot-untangling user study ($N$ = 22) using ropes with varying length and rigidity. 
Results show that the effectiveness of the assistance depends strongly on operator expertise and DLO properties. 
SA-CBF provides the strongest gains for naive users, acting as a skill equalizer that increases task success from 71\% to 88\%, and is effective for stiffer ropes. Conversely, expert users prefer VA, and highly compliant, long ropes benefit more from visual support than localized action assistance. 
Ultimately, these findings demonstrate that effective DLO teleoperation cannot rely on a fixed strategy, highlighting the critical need for adaptive, user-aware, and material-aware shared autonomy.
\end{abstract}

% Notice that Elsevier uses \begin{keyword} and separates items with \sep
\vspace{1em}
\vspace{2em}

%\end{frontmatter}

\section{Introduction}
\label{sec:intro}

\begin{figure}[t!]
    \centering
    \includegraphics[width=\linewidth]{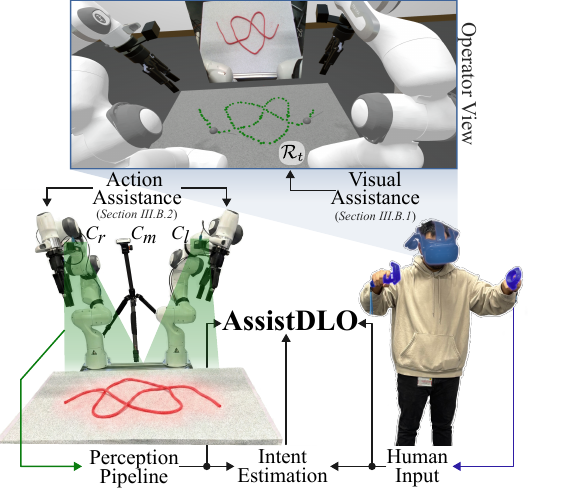}
    \vspace{-1em}
    \caption{System overview of \textit{AssistDLO}. Dual wrist-mounted RGB-D cameras ($C_l$ and $C_r$) continuously estimate the DLO state $\mathcal{R}_t$ in real time. By combining human input with $\mathcal{R}_t$ to estimate the operator's intention, \textit{AssistDLO} provides two forms of support: \hyperref[VA]{Visual Assistance (VA)} to augment perception of the human operator through visual cues $\mathcal{R}_t$, and Action Assistance via our novel control barrier function (CBF)-based shared autonomy controller (\hyperref[CBF]{SA-CBF}) for geometry-aware DLO grasping.}
    \label{fig:main}
\end{figure}

In industrial automation, tasks such as automotive wire harness assembly and aircraft fuselage wiring remain heavily dependent on manual labor~\cite{sanchezRoboticManipulationSensing2018,salunkhe2023specifying}. In the aerospace sector, the autonomous robotic outfitting of lunar surface cables and the management of flexible tethers are critical for advanced structural assembly~\cite{quartaroModelingDeformableLinear2024,shahAircraftAssembly2018,adebolaAutomatingDeformableGasket2024}.
Similarly, in the medical domain, the precise routing of sutures by surgical robots is fundamental to cardiothoracic procedures~\cite{kitagawaEffectSensorySubstitution2005,barba2022remote}. 
These diverse, high-value applications all share a common core challenge, which is the robotic manipulation of Deformable Linear Objects (DLOs).

Fully autonomous solutions have made significant progress in specialized tasks, including shape control~\cite{aksoyPlanningControlDeformable2026,almaghoutRoboticComanipulationDeformable2024,huangLearningGraphDynamics2023}, cable routing~\cite{chenContactAwareShapingMaintenance2023,chenMultiRobotAssemblyDeformable2025,zhangHarnessingTwistingSingleArm2024}, and knot untangling~\cite{yuHANDLOOM30Interactive, dinkelKnotDLOInterpretableKnot, grannenUntanglingDenseKnots2020, shivakumarSGTM20Autonomously2022, viswanathHANDLOOMLearnedTracing2023a}. 
However, DLOs possess an infinite-dimensional configuration space and exhibit highly non-linear dynamics, making accurate analytical modeling computationally expensive~\cite{yinModelingLearningPerception2021}. 
Because they are inherently underactuated, minor end-effector motions can induce large, unpredictable global deformations. 
Consequently, autonomous planning and control algorithms remain highly vulnerable to failure when faced with unmodeled physical interactions or complex self-intersections~\cite{zhangHarnessingTwistingSingleArm2024,aksoyCollaborativeManipulationDeformable2024,benschPhysicsInformedNeuralNetworks2024,yuGlobalModelLearning2023}. 
When autonomous recovery cannot be guaranteed, or when high-level planning is infeasible across the entire configuration space, teleoperation remains a critical approach for ensuring safe execution in unstructured environments~\cite{chiaravalliVisionbasedSharedAutonomy2023}. 

While teleoperation provides flexibility for online adaptation during robotic manipulation and recovery from execution failures, it also requires significant physical and mental effort from the human operator.
For instance, teleoperation inherently suffers from embodiment mismatch and degraded depth perception when operators rely on 2D camera feeds~\cite{su_mixed_2022, hebriTeleoperationFrameworkRobots2023a, linPerceptionActionAugmentation2024}. 
These difficulties are further amplified in DLO manipulation, where establishing precise contact with thin, featureless ropes is particularly challenging~\cite{zhu2021challengesoutlookroboticmanipulation}. 
In addition, the need for continuous bimanual coordination to maintain DLO shape increases the operator's cognitive load. 
Furthermore, if a stable grasp is lost, an unconstrained DLO may rapidly deform into a complex new configuration~\cite{sanchezRoboticManipulationSensing2018,aksoyCollaborativeManipulationDeformable2024}. 
In tasks like knot untangling, such uncontrolled deformations can unintentionally tighten existing loops or move critical segments out of the camera's view or the robot's reachable workspace, rapidly leading to task failure~\cite{shivakumarSGTM20Autonomously2022, matsunoManipulationDeformableLinear2006, viswanathHANDLOOMLearnedTracing2023a}.

To mitigate these challenges, researchers have developed assistive teleoperation strategies, broadly categorized into perception assistance (enhancing situational awareness)~\cite{linPerceptionActionAugmentation2024, krishnanHumanPreferredAugmented2023} and action assistance (blending human intent with robotic precision)~\cite{draganFormalizingAssistiveTeleoperation2013, gottardiSharedControlRobot2022}. 
Traditional shared autonomy frameworks frequently use linear policy blending, interpolating between the human's input and a computed autonomous grasp pose~\cite{muellingAutonomyInfusedTeleoperation2015, gopinathHumanintheLoopOptimizationShared2017, gottardiSharedControlRobot2022, javdaniSharedAutonomyHindsight2017}. 
While highly effective for rigid-object manipulation, simple linear blending acts as a context-agnostic ``geometric magnet''. 
In the context of DLO manipulation, this behavior can be particularly harmful because the end-effector may fail to reach the intended target smoothly and instead displace ungrasped rope parts, altering the topology and worsening the entanglement before a secure grasp is established.

While there are existing studies on DLO manipulation via teleoperation, the majority focus on Learning from Demonstration (LfD) to extract offline policies~\cite{leeLearningForcebasedManipulation2015, rambowAutonomousManipulationDeformable2012, liGRRLGoingDexterous2025}. 
Research dedicated to real-time assisted teleoperation for DLOs remains limited~\cite{grc, chiaravalliVisionbasedSharedAutonomy2023}. 
Consequently, there is a lack of analysis of which forms of assistance are most effective for different levels of human operator expertise and for varying physical properties of DLOs, even though this information can be important for critical DLO manipulations in industry, space, or medical domains.

To address this gap, we introduce \textit{AssistDLO}, a comprehensive bimanual assistive teleoperation framework for DLO manipulation. 
Driven by a real-time multi-view perception pipeline, we propose a novel context-aware shared autonomy controller (SA-CBF). 
Unlike traditional blending methods, SA-CBF uses Control Barrier Functions (CBF) to construct a dynamic, geometry-aware ``safety funnel'' that enables precise grasping without overriding the operator's high-level intent or disrupting the surrounding rope topology.

To systematically evaluate the effectiveness of \textit{AssistDLO}, we consider bimanual knot untangling, a highly unconstrained task that involves complex DLO deformations, including self-intersections, and topology-sensitive grasping. 
We conducted a comprehensive user study ($N=22$) to evaluate performance across varying DLO properties, including flexural rigidity, total mass, and length, as well as across different levels of human operator expertise.
In summary, this study makes the following contributions:

    \noindent \textbf{Assistive Teleoperation Framework:} A bimanual assistive teleoperation framework for DLO manipulation that integrates multi-view DLO state estimation, intent estimation, visual assistance, and action assistance.
    
    \noindent \textbf{Context-Aware Action Assistance:} A CBF-based local guidance mechanism (SA-CBF) that supports geometry-sensitive grasping while preserving operator authority, mitigating the unintended DLO displacements often introduced by standard linear blending.
    
    \noindent \textbf{Empirical Human-Subject Evaluation:} A systematic user study that quantitatively and qualitatively characterizes how the effectiveness of teleoperation assistance depends on operator expertise and physical properties of the DLO.
    
    \noindent \textbf{Design Insights for DLO Shared Autonomy:} Evidence-based insights showing that action assistance is particularly beneficial for naive users and for stiffer ropes, whereas expert users and highly compliant materials benefit more from visual assistance.

\section{Related Work} \label{sec: RW}

The scope of this work encompasses the intersection of  \hyperref[subsec:RW_DLOMANIP]{DLO manipulation} and \hyperref[subsec:RW_ASSISTELEOP]{assistive teleoperation}. 
Existing work largely studies DLO perception, autonomous DLO manipulation, or teleoperation interfaces in isolation. 
This section reviews the foundations in these domains to contextualize the need for task-aware assistance in unconstrained DLO manipulation.

\subsection{DLO Manipulation and Perception} \label{subsec:RW_DLOMANIP}

In autonomous DLO manipulation, perception is typically required to extract the object's topological state or key features from sensor data, while modeling is utilized for forward simulation to predict the object's deformation in response to planned control actions~\cite{wangOfflineOnlineLearningDeformation2022, yuGlobalModelLearning2023, yanSelfSupervisedLearningState2020, tangLearningBasedMPCSafety2024, zhangParticleGridNeuralDynamics2025}. 
Although recent surveys highlight significant advancements in both perception and dynamics modeling~\cite{zhu2021challengesoutlookroboticmanipulation, yinModelingLearningPerception2021, sanchezRoboticManipulationSensing2018}, unconstrained tasks involving large deformations, such as knot untangling, remain highly challenging~\cite{viswanathHANDLOOMLearnedTracing2023a, shivakumarSGTM20Autonomously2022}. 
In these scenarios, current perception and modeling methods struggle with high occlusion, complex deformations, and non-linear dynamics.

Analytical DLO modeling methods, such as multi-body mass-spring-damper systems, continuous Cosserat rod formulations, and Position-Based Dynamics, rely on explicit physical equations to simulate DLO behavior~\cite{provotDeformationConstraintsMass, quartaroModelingDeformableLinear2024, caporaliRoboticManipulationDeformable2025, muellerPBD2007}. 
Although these methods provide structured physical constraints and have been extended to capture complex deformations, these improvements often come with an increased computational cost.
This trade-off can limit their use in computationally limited control and planning frameworks~\cite{longSelfsupervisedPhysicsInformedManipulation2026, aksoyCollaborativeManipulationDeformable2024}.
In addition, their accuracy can still degrade under large deformations, and they typically require manual calibration of physical parameters, such as stiffness, damping, and compliance, to adapt to uncontrolled real-world environments~\cite{quartaroModelingDeformableLinear2024}.

To address computational bottlenecks and parameter tuning of analytical frameworks, recent research has focused on data-driven modeling. 
Graph Neural Networks (GNNs) and self-supervised dynamics models learn DLO behaviors directly from simulation data, accommodating severe self-intersections~\cite{wangOfflineOnlineLearningDeformation2022, yanSelfSupervisedLearningState2020}. 
In collaborative settings, approaches such as topological latent control models reactively handle DLO deformations caused by external forces~\cite{zhou_reactive_2024}. 
However, learning-based modeling approaches inherently struggle with out-of-distribution generalization, frequently failing when deployed in DLOs with unseen physical properties or topological states~\cite{caporaliDeformableLinearObjects2024, aksoyCollaborativeManipulationDeformable2024, longSelfsupervisedPhysicsInformedManipulation2026}.

Due to the high topological complexity of tasks such as knot untangling, accurate dynamics modeling is often a computational bottleneck in real-time. 
Consequently, recent frameworks focus on model-free policies that bypass modeling. 
For instance, HULK uses neural networks to directly learn key points relevant to the task from images, such as optimal grasping and pulling points, without simulating the DLOs' dynamics~\cite{grannenUntanglingDenseKnots2020}. 
Similarly, others bypass explicit dynamics by learning recovery policies and perception features to iteratively untangle dense, non-planar knots~\cite{sundaresanUntanglingDenseNonPlanar2021, viswanathHANDLOOMLearnedTracing2023a}. 
To handle longer, highly entangled cables, systems like SGTM 2.0 and HANDLOOM employ interactive perception and autoregressive tracing to sequentially estimate the cable's spatial state~\cite{shivakumarSGTM20Autonomously2022, viswanathHANDLOOMLearnedTracing2023a}. 
Despite advances such as bidirectional tracing~\cite{yuHANDLOOM30Interactive}, perception in cluttered cable arrangements remains challenging due to variations in the physical properties of DLOs. 
In addition, closely spaced thin cables are difficult to disambiguate under the fixed depth perception of a monocular overhead camera. Given the shift away from explicit dynamics in unconstrained tasks, reliable perception and state estimation emerge as the critical bottlenecks. Errors in state estimation, such as misinterpreting severe self-intersections, directly propagate to control failures.

Existing perception pipelines span multiple subproblems, including segmentation, 2D shape estimation, 3D shape estimation, and tracking~\cite{xiangTrackDLOTrackingDeformable2023, caporaliRTDLORealTimeDeformable2023, luoTSLTrackingDeformable2025, caporaliDeformableLinearObjects2023, kickiDLOFTBsFastTracking2023}. Methods such as DLOFTBs focus on 2D shape estimation from image observations, whereas approaches such as TrackDLO address temporally consistent tracking of DLO state. For instance, TrackDLO leverages geometric and topological priors from a monocular view to achieve robust real-time tracking of a single DLO under partial occlusion, but it requires accurate initialization and may degrade after tracking failures~\cite{xiangTrackDLOTrackingDeformable2023}. In contrast, DLOFTBs removes the need for explicit initialization through precomputed binary masks, yet it remains limited to 2D shape estimation and monocular fixed-view perception, which makes severe 3D depth ambiguities at dense knot crossings difficult to resolve~\cite{kickiDLOFTBsFastTracking2023}. To address such ambiguities, multi-view observations can be incorporated, but existing approaches still rely on a model-based reconstruction pipeline, whose generalization across different DLO properties remains limited~\cite{caporaliRoboticManipulationDeformable2025}.

For DLO segmentation, more recent studies have explored promptable foundation models such as SAM 2, which offer strong zero-shot performance in cluttered scenes~\cite{raviSAM2Segment2024, zhaoleRobustDeformableLinear2024}. 
However, these models still require careful prompting to disambiguate the target instance, and DLO-specific pipelines often augment them with bounding boxes, text embeddings, or tailored prompts to improve robustness under occlusion~\cite{zhaoleRobustDeformableLinear2024, kozlovskyISCUTEInstanceSegmentation2024}. 
Nevertheless, segmentation alone is insufficient, as pixel-level masks still leave ambiguities in the underlying 3D structure and self-intersection, particularly in single-view settings.

\subsection{Assistive Teleoperation} \label{subsec:RW_ASSISTELEOP}

Assistive teleoperation aims to support a human operator in the execution of a given task.  
AssistDLO provides two distinct levels of assistance: \hyperref[subsubsec:RW_ASSISTELEOP_ACTION]{Action Assistance}, which employs shared autonomy to facilitate the human operator in grasping the DLO, and \hyperref[subsubsec:RW_ASSISTELEOP_VISUAL]{Visual Assistance}, which enhances the operator's situational awareness by providing visual aids.

\subsubsection{Action Assistance} \label{subsubsec:RW_ASSISTELEOP_ACTION}

Assistive teleoperation aims to reduce operator workload and improve task success in complex scenarios by fusing human input with autonomous assistance~\cite{selvaggioAutonomyPhysicalHumanRobot2021}. 
Importantly, the majority of shared autonomy methods in the literature focus almost exclusively on rigid object manipulation.
A fundamental challenge in this domain is balancing autonomous intervention with the operator's sense of agency, requiring accurate intent estimation~\cite{draganTeleoperationIntelligentCustomizable2013, manschitzSharedAutonomyIntuitive2022}. Approaches in the literature infer intent based on a finite set of goals~\cite{gottardiSharedControlRobot2022}, dynamic trajectory tracking~\cite{hauserRecognitionPredictionPlanning2013}, or gaze-tracking~\cite{fuchsGazeBasedIntentionEstimation2021}. 
Methods often employ Recursive Bayesian Filtering or Hidden Markov Models to decode sequential observations~\cite{aronsonInferringGoalsGaze2021}. 
However, extracting human intention specifically for DLO manipulation during teleoperation remains largely unexplored.

Once intent is estimated, shared autonomy systems harmonize user commands with autonomous assistance. The foundational paradigm is linear policy blending, where the executed action is a weighted combination of inputs~\cite{draganPolicyblendingFormalismShared2013}. Other frameworks model shared control as a Partially Observable Markov Decision Process (POMDP) to optimize goal-aware assistance policies~\cite{javdaniSharedAutonomyHindsight2018}. More recently, vision-based shared autonomy frameworks have been explored for manipulation with unknown and dynamically changing rigid-object locations~\cite{belsareZeroShotUserIntent2025}. Learning-based assistance provides additional flexibility by implicitly inferring intent through methods such as deep reinforcement learning and diffusion-based policies~\cite{reddySharedAutonomyDeep2018, yonedaNoiseBackDiffusion2023}. However, safety and interpretability remain central concerns. Related work has also explored sampling-based shared autonomy, where robot commands are generated to satisfy safety constraints over sampled configurations~\cite{manschitzSamplingBasedGraspCollision2025}. 

To provide formal safety guarantees, recent shared autonomy frameworks have incorporated Control Barrier Functions (CBFs)~\cite{ames2019cbf}. CBF-based filters have been used to constrain interaction forces in robotic surgery~\cite{qinHapticSharedControl2025} and to ensure geometric safety by filtering user commands at the kinematic level to avoid obstacles~\cite{zhang2021cbfteleop, berk2026barrierik}. Other work has incorporated human cognitive states directly into the CBF formulation as real-time inequality constraints~\cite{uzunArbitrationControlBarrier2025}.

Literature specifically targeting shared autonomy for DLO manipulation remains limited. One representative contribution introduces a vision-based shared autonomy framework that enables the human operator to alternate between unconstrained teleoperation and a cable-targeting operational mode~\cite{chiaravalliVisionbasedSharedAutonomy2023}. When targeting is activated, virtual fixtures constrain the manipulator to follow the longitudinal axis of the detected cable, thereby reducing the need for precise manual alignment. However, this approach focuses on relatively simple cable-routing tasks rather than the severe topological complexities encountered in knot untangling.

\subsubsection{Visual Assistance} \label{subsubsec:RW_ASSISTELEOP_VISUAL}

While action assistance modulates physical control commands, visual assistance addresses the cognitive and sensory bottlenecks that the human operator experiences. On the remote side, operators often suffer from a loss of situational awareness, as standard 2D camera feeds obscure depth perception and distort spatial relationships. Augmented Reality (AR), Virtual Reality (VR), and Mixed Reality (MR) interfaces have emerged as critical tools for visually augmenting the workspace~\cite{livatino2021MRVA, linPerceptionActionAugmentation2024}.

Several methods overlay abstract sensory information directly onto the operator's visual field. To reduce cognitive workload, target locators and collision alerts can be displayed over a live video stream~\cite{linPerceptionActionAugmentation2024}, while other approaches render spatial cues and ``ghost'' objects to visualize future robot poses~\cite{arevaloarboledaAssistingManipulationGrasping2021}. For tasks requiring high-dimensional awareness, frameworks utilizing Point Cloud Augmented Virtual Reality map live visual data onto wider synthetic 3D contexts to guide the operator safely~\cite{niPointCloudAugmented2018}. In robotic teleoperation, AR and MR frameworks leverage RGB-D imaging to merge 3D and 2D views, providing essential depth cues for intuitive manipulation~\cite{pan_augmented_2021, su_mixed_2022}. Recently, MR has been explored for human-to-robot skill transfer in contact-rich assembly tasks~\cite{wu_mixed_2026}, though these interfaces do not specifically address the perceptual difficulties of revealing DLO topologies.

\begin{comment}
\subsection{Gap and Positioning} \label{subsec:RW_GAP}

Prior studies typically identifies limitations in state-of-the-art DLO tracking to justify novel autonomous algorithms, or it optimizes teleoperation metrics under the assumption of rigid environments.  In contrast, our focus is on how different assistance modes interact with operator expertise and DLO properties in an online manipulation setting.

To bridge this gap, our proposed \textit{AssistDLO} framework completely bypasses explicit dynamics modeling and global autonomous state reconstruction. Instead, it assumes that human cognition is necessary for untangling dense topologies. We introduce a robust, model-free visual perception pipeline tailored specifically to feed a 3D VR interface that augments the user's view with a real-time 3D representation of the DLO, resolving depth ambiguities for the human operator. Concurrently, we replace standard linear blending which is highly vulnerable to altering and tightening DLO topologies with a novel, topology-aware shared autonomy mechanism based on Control Barrier Functions. By combining these visual and action-based assitance modalities, we target the specific cognitive and haptic requirements of unconstrained DLO teleoperation.
\end{comment}
\section{AssistDLO} \label{sec:AssistDLO}

Motivated by the gaps identified in \hyperref[sec: RW]{Section~\ref*{sec: RW}}, the \textit{AssistDLO} framework addresses the perceptual and control challenges of unconstrained DLO manipulation without relying on explicit dynamics modeling or full 3D reconstruction of the DLO state. 
Instead, the framework is guided by the principle that human cognition remains essential for reasoning about complex deformations and high-density topological configurations, and that the robotic system should be designed to augment, rather than replace, the operator's manipulation and perception capabilities. 

To provide real-time assistance, AssistDLO is built around two complementary components. For the first component, we introduce a model-free visual perception pipeline. Notably, while multi-view state estimation remains relatively uncommon in the DLO literature, our approach utilizes dual-camera inputs to feed both the 3D VR interface and the downstream control modules. 
For the second component, we address a gap in the literature by proposing a novel, geometry-aware shared-autonomy method tailored specifically to DLOs. By combining these visual and action-based modalities, \textit{AssistDLO} directly targets the specific cognitive and control requirements of unconstrained DLO teleoperation.

\subsection{DLO State Estimation}
\begin{figure}[t]
    \centering
    \includegraphics[width=1\linewidth]{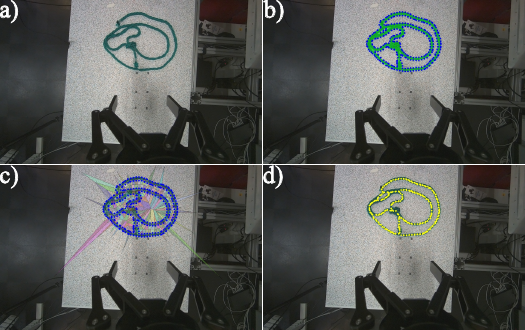}
    \vspace{-1em} \caption{Overview of the proposed DLO trace extraction pipeline from an RGB view captured from a single wrist camera. (a) Original image $I_t$. (b) Refined rope segmentation $M_t$ shown as a green overlay, with sparse contour samples $C_t$ (blue dots). (c) Voronoi diagram computed from the contour samples (cells between edges rendered with low transparency with different colors) to reveal candidate medial-axis structures within the mask. (d) Resulting vertices (yellow dots) in the pruned graph $G_t$ after removing vertices outside the rope mask, yielding a clean set of rope trace points for subsequent 3D back-projection and fusion.}
    \label{fig:DLOStateEstimation}
\end{figure}
In the context of this study, DLO state estimation refers specifically to the task of detecting a single rope and continuously inferring its 3D geometric pose from multi-view RGB-D data in real time. Our objective is not full physical reconstruction or parameter identification of the rope, but to get a task-relevant geometric estimate that supports local grasp guidance and visual augmentation. We represent the DLO state non-parametrically as a dense sequence of 3D spatial points, focusing entirely on observable spatial configurations rather than inferring intrinsic dynamic parameters such as stiffness, friction, or mass. This level of representation is sufficient for the assistive functions studied in this paper, which depend primarily on local geometric consistency near grasping and manipulation regions.

Many state-of-the-art DLO state estimation algorithms rely on initializations with no self-intersections and single-view, fixed-camera setups. These assumptions fundamentally limit their ability to resolve 3D depth ambiguities at dense knot crossings.

The inherent physical and geometric characteristics of DLOs make their tracking a fundamentally challenging problem. In particular, DLOs exhibit high sensitivity to even minimal external forces, which induce rapid, non-rigid shape transformations (deformations). Second, due to their small thickness, depth measurements are often noisy, degrading 3D state estimation. Third, surface textures and reflections can reduce the reliability of color-based segmentation. Finally, in teleoperation setups, the environment is often uncontrolled, and the background may contain artifacts that degrade visual perception.

To address these challenges, our perception pipeline, depicted at a high level in \hyperref[fig:DLOStateEstimation]{Figure~\ref*{fig:DLOStateEstimation}}, consists of two primary stages: (i) 2D DLO trace extraction from individual camera streams, and (ii) multi-view fusion to estimate the full 3D spatial state. For this pipeline, we use left ($C_l$) and right ($C_r$) wrist-mounted RGB-D cameras, as shown in \hyperref[fig:main]{Figure~\ref*{fig:main}}. At time $t$, each camera captures an RGB image $I_t \in \mathbb{R}^{H \times W \times 3}$ and a depth map $D_t \in \mathbb{R}^{H \times W}$, denoted as ($I_t^l, D_t^l$) and ($I_t^r, D_t^r$), respectively. 

Specifically, we adopt the baseline trace extraction method proposed in~\cite{grc}. We build upon this baseline by dynamically fusing observations from two wrist-mounted cameras to actively resolve depth ambiguities, rather than fusing multiple possible grasping poses. Furthermore, instead of relying on mathematical global 3D topological assumptions, our approach derives the topological state directly from segmented masks via Voronoi-based skeletonization, coupled with graph networks for structural pruning. This enables reliable, real-time state extraction without relying on strict environmental assumptions.

\begin{figure}[t]
    \centering
    \includegraphics[width=1\linewidth]{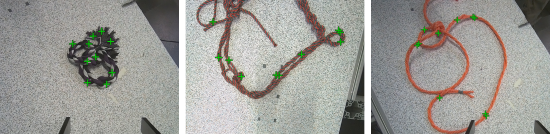}
    \vspace{-1em} \caption{Representative samples from the labeled dataset, featuring four distinct rope textures and properties and 41 images annotated with positive point prompts (indicated with green plus markers).}
    \label{fig:dataset}
\end{figure}

\textbf{DLO Trace Extraction.} Each camera independently executes the trace extraction module to isolate the DLO's geometry in the image plane. The primary objective of this initial step is to obtain a robust 2D binary mask of the rope. Representative HSV segmentation outputs from the wrist camera are provided in the supplementary video to illustrate the limitations of standalone color thresholding in this environment. To overcome these limitations, we employ SAM~2~\cite{raviSAM2Segment2024} from among the various advanced instance segmentation models available. To initialize the model at $t=0$, we provide a small set of manually labeled point prompts on reference images, as shown in \hyperref[fig:dataset]{Figure~\ref*{fig:dataset}}. While other initialization methods for SAM~2 could be achieved by coupling SAM~2 with vision-language models, for instance, by scoring all generated segments using image-text models with prompts such as ``rope'' or ``cable''~\cite{schuhmann2022laionb}, we opted for a different approach. Instead, we chose to initialize using a labeled dataset to ensure high reliability for the dependent assistance modules. Although a formal ablation of initialization techniques is beyond the scope of this work, empirical observations confirm that explicitly validated initial prompts minimize early-stage segmentation errors. 

For all subsequent frames, the model operates in continuous tracking mode to generate a primary spatial mask $S(I_t)$. While this foundational model provides robust zero-shot tracking, it can occasionally over-segment the tight inner loops of dense knots, mistakenly including enclosed background pixels. To enforce strict geometric boundaries, we compute an auxiliary mask $B(I_t)$ by converting $I_t$ to the HSV color space and applying Otsu thresholding~\cite{otsu}. The final refined DLO mask $M_t = S(I_t) \wedge B(I_t)$ is obtained via the pixel-wise logical AND ($\wedge$) operation, as shown in \hyperref[fig:DLOStateEstimation]{Figure~\ref*{fig:DLOStateEstimation}b}. 

From $M_t$, we extract a sparse contour set $C_t$ (see blue dots in \hyperref[fig:DLOStateEstimation]{Figure~\ref*{fig:DLOStateEstimation}c}) using grid search methods in the image plane. We then compute the Voronoi diagram $(V_t,E_t)=\mathrm{Voronoi}(C_t)$, which yields a set of vertices $V_t$ and edges $E_t$. Using $V_t$ and $E_t$, we construct a graph $G_t=(V_t,E_t)$. The primary objective of this step is to extract a clean, 1D skeletal graph that represents the DLO's geometry. However, the raw Voronoi graph initially contains false positives—such as edges and vertices that extend beyond the physical boundaries of the rope—which must be filtered out. Concretely, we prune the graph such that a vertex is retained only if it falls within $M_t$, and an edge is retained only if its endpoints and uniformly sampled interior points all lie strictly within $M_t$. In the resulting pruned graph $\hat{G}_t$, the degree of a node is defined as its number of connected edges, and terminal nodes are vertices with a degree of exactly one. Because the Voronoi-based method inherently generates artificial short branches, the graph may still contain more than two terminal nodes. To resolve this, we identify the endpoints of the primary visible trace as the node pair separated by the maximum geodesic distance within the graph. We then iteratively prune the remaining shorter terminal branches to finally extract a single, continuous DLO trace.

Each $i^{\text{th}}$ surviving node in the pruned graph $\hat{G}_t$ corresponds to a pixel coordinate $\mathbf{u}_{i,t}=[u_{i,t}\ v_{i,t}]^\top$. Given the corresponding depth measurement $z_{i,t}=D_t(\mathbf{u}_{i,t})$ and the camera intrinsic parameters $(f_x,f_y,c_x,c_y)$, the 3D position of the node in the camera frame is obtained via pinhole back-projection as
\[
\mathbf{p}_{i,t}=
\begin{bmatrix}
(u_{i,t}-c_x)\cdot z_{i,t}/f_x\\
(v_{i,t}-c_y)\cdot z_{i,t}/f_y\\
z_{i,t}
\end{bmatrix},
\]
which yields the local spatial coordinates required for the subsequent multi-view fusion.

\textbf{Multi-View DLO State Fusion and Estimation.} 
Following trace extraction, we obtain the 3D point sets representing the DLO in the local frames of the left $C_l$ and right $C_r$ cameras. We denote these local point sets as $\mathcal{P}^{(l)}_t$ and $\mathcal{P}^{(r)}_t$, respectively. Since the cameras are wrist-mounted, their extrinsic parameters are continuously updated using the robot's forward kinematics, as shown in \hyperref[fig:main]{Figure \ref*{fig:main}}. Each point is mapped into the common task frame $\{O\}$ as
\[
\begin{bmatrix}
{}^{O}\mathbf{p}^{(c)}_{i,t} \\ 1
\end{bmatrix} 
= {}^{O}\mathbf{T}_{C_c}(t)
\begin{bmatrix}
\mathbf{p}^{(c)}_{i,t} \\ 1
\end{bmatrix},
\]
using the homogeneous transformation ${}^{O}\mathbf{T}_{C_c}(t) \in SE(3)$ for camera $c\in\{l,r\}$. 

Having multiple views of the DLO defined in the same coordinate frame allows us to directly combine this spatial information. To achieve this, we first mitigate sensory latency by evaluating each camera's observation independently; any point cloud older than a timeout threshold (i.e., $1.0$\,s) is discarded prior to fusion. Subsequently, we actively aim to resolve single-view depth ambiguities and severe occlusions by fusing the valid transformed point sets via a union operation, yielding the combined global set of 3D points
\[
\mathcal{R}_t = \mathcal{P}^{(l)}_{t,O} \cup \mathcal{P}^{(r)}_{t,O}.
\]
This fused 3D DLO point set $\mathcal{R}_t$ serves as the non-parametric state representation of the DLO.

Because this raw fused point set $\mathcal{R}_t$ contains depth quantization noise and highly dense overlapping regions from the dual-camera setup, we spatially regularize the data using voxel-grid downsampling~\cite{Rusu_ICRA2011_PCL}. 
To simultaneously satisfy the differing requirements of the assistive modalities, we implement a dual-resolution filtering scheme. \label{quote:downsample}
The role of the perception pipeline in this study is therefore to provide a reliable online geometric foundation for the downstream teleoperation interface, rather than to propose a standalone state-estimation algorithm.

\subsection{Assisted Teleoperation}
Precise grasping of thin, deformable objects is notoriously difficult in standard teleoperation due to inherent depth-perception errors and the mismatch between human and robotic embodiment. Leveraging the robust, non-parametric DLO state representation $\mathcal{R}_t$ obtained via multi-view fusion, our framework actively aims to mitigate these cognitive and physical demands by providing two complementary layers of support: (i) visual assistance to enhance the operator's depth cognition, and (ii) action assistance to dynamically augment their continuous control inputs. 

\subsubsection{Visual Assistance and Teleoperation Framework}\label{VA}
In our teleoperation framework, the human operator is immersed in a virtual reality (VR) environment, utilizing a Head-Mounted Display (HMD) and tracked hand controllers to command the continuous 6-DoF task-space poses of the bimanual robot end-effectors in real-time. To minimize network bandwidth and maintain the high frame rates (FPS) required to prevent operator motion sickness, the baseline visual feedback from the remote workspace is limited to a compressed 2D RGB video stream captured by a central main camera ($C_m$), as shown in \hyperref[fig:main]{Figure \ref*{fig:main}}. The operator can freely move their head within the local 3D VR environment. However, the flat 2D projection of the remote scene inherently limits the depth information necessary for precise DLO grasping.

Rather than treating augmented visual feedback as a simple baseline, we formulate Visual Assistance (VA) as a distinct, primary assistance strategy. Visual assistance targets perceptual uncertainty rather than motor execution, making it especially suitable for operators who already possess strong control priors but benefit from enhanced scene understanding.

To bridge the perceptual gap without violating bandwidth constraints, the VA modality actively provides a wider spatial context by augmenting the operator's interface with lightweight 3D spatial feedback. Rather than attempting to stream dense, computationally heavy point clouds of the entire remote workspace, we transmit only the compact, regularized DLO state representation $\mathcal{R}_t$. Specifically, we utilize the finer auxiliary point cloud (downsampled as mentioned in \ref{quote:downsample}) to preserve visual topology while keeping bilateral communication lightweight. By localizing the complex rope structure, this approach provides non-intrusive assistance that preserves the operator's full manual control.

The localized 3D point set $\mathcal{R}_t$ is rendered dynamically within the operator's virtual environment (green points depicted in operator view in \hyperref[fig:main]{Figure \ref*{fig:main}}), spatially registered to align with the 2D RGB stream. Because the operator views this augmented scene through an HMD, the rendered DLO state naturally provides stereoscopic depth cues. Furthermore, as the user moves their head, motion parallax offers immediate, intuitive geometric feedback regarding the exact 3D spatial relationship between the robot grippers and the rope, significantly reducing depth ambiguity and the cognitive burden of depth estimation.

\begin{algorithm}[t]
\begin{algorithmic}[1]
\Require 3D DLO point set~$\mathcal{R}_t$, human target~$\mathbf{x}^{h,j}_t$ and robot pose~$\mathbf{x}^{r,j}_t$ for each arm $j \in \{l,r\}$, Previous target~$\hat{\mathbf{p}}^j_{t-1}$ (if available), step radius~$r_{\mathrm{step}}$, robot proximity threshold~$\epsilon_{\mathrm{robot}}$
\Ensure Autonomous target pose $\mathbf{u}_a^j = (\mathbf{x}^{a,j}_t, \mathbf{q}^{a,j}_t)$

\State \textit{// 1. Global Nearest Neighbor Search}
\State $\mathbf{p}^* \gets \arg\min_{\mathbf{p} \in \mathcal{R}_t} \|\mathbf{p} - \mathbf{x}^{h,j}_t\|_1$

\State \textit{// 2. Topological Continuity \& Hysteresis}
\If{$\hat{\mathbf{p}}^j_{t-1}$ is valid \textbf{and} $\|\mathbf{x}^{r,j}_t - \hat{\mathbf{p}}^j_{t-1}\|_2 < \epsilon_{\mathrm{robot}}$}
    \State \textit{// Robot is engaged; restrict to local neighborhood $\Omega_t$ }
    \State $\Omega_t \gets \{\mathbf{p} \in \mathcal{R}_t \mid \|\mathbf{p} - \hat{\mathbf{p}}^j_{t-1}\|_2 \le r_{\mathrm{step}}\}$
    \State $\hat{\mathbf{p}}^j_t \gets \arg\min_{\mathbf{p} \in \Omega_t} \|\mathbf{p} - \mathbf{x}^{h,j}_t\|_1$
\Else  \textit{  // No previous target exists}
    \State $\hat{\mathbf{p}}^j_t \gets \mathbf{p}^*$
\EndIf

\State \textit{// 3. Tangent Estimation \& Top-Down Orientation}
\State $\mathbf{t}_t \gets \mathrm{PCA}(\mathrm{kNearestNeighbors}(\hat{\mathbf{p}}^j_t, \mathcal{R}_t))$
\State $\psi_t \gets \mathrm{atan2}(\mathbf{t}_{t,y}, \mathbf{t}_{t,x})$ \textit{//Project tangent to horizontal plane}
\State $\mathbf{q}^{a,j}_t \gets \mathrm{EulerToQuaternion}(0, 0, \psi_t)$

\State \textit{// 4. Final Autonomous Target Pose}
\State $\mathbf{x}^{a,j}_t \gets \hat{\mathbf{p}}^j_t$
\State $\mathbf{u}_a^j \gets (\mathbf{x}^{a,j}_t, \mathbf{q}^{a,j}_t)$

\State \Return $\mathbf{u}_a^j$
\end{algorithmic}
\caption{Human Intent Estimation and Target Selection}
\label{alg:intent_estimation}

\end{algorithm}
\subsubsection{Action Assistance}\label{AA}
%aksoyun maili bir tane daha vardi ya

The human operator's continuous control input is denoted as $\mathbf{u}_h=(\mathbf{u}_h^l,\mathbf{u}_h^r)$ for the left and right robot arms in the bimanual setup. For each arm $j \in \{l,r\}$, the commanded end-effector target pose in the task space is $\mathbf{u}_h^j \in SE(3)$, which we decompose as $\mathbf{u}_h^j=(\mathbf{x}^{h,j}_t,\mathbf{q}^{h,j}_t)$, where $\mathbf{x}^{h,j}_t\in\mathbb{R}^3$ denotes the target position and $\mathbf{q}^{h,j}_t\in\mathbb{H}$ is the target orientation expressed in quaternion form. In the unassisted baseline, these human commands are continuously used as final target poses $\mathbf{u}_f=(\mathbf{u}_f^l, \mathbf{u}_f^r)$ in task space, $\mathbf{u}_f = \mathbf{u}_h$, tracked by an inverse kinematics (IK) solver equipped with standard null-space projections for real-time self-collision avoidance.

\textbf{Human Intent Estimation.} Assistance is only useful when applied near the operator's intended interaction region. To provide effective action assistance, the system must first infer the operator's intended grasp location. Unlike rigid object manipulation, where discrete grasp affordances (e.g., a mug handle) can be predefined, a DLO presents a continuous manifold of potential grasp points. Furthermore, human input trajectories are inherently free-form; operators do not follow mathematically optimal paths to a grasp. Therefore, we treat every regularized node in the downsampled 3D point set $\mathcal{R}_t$ as a valid candidate for grasping. \label{sec:hie}

\begin{figure}[t]
    \centering
    \includegraphics[width=1\linewidth]{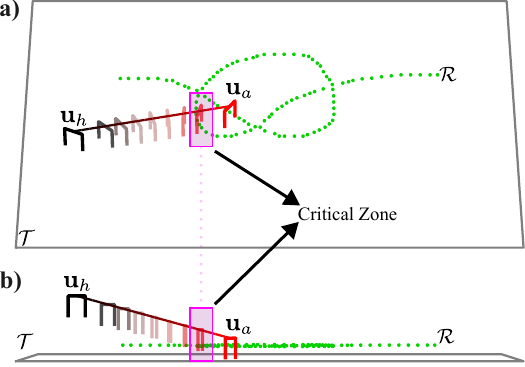}
    \vspace{-1em}
    \caption{\textbf{Shared Autonomy via Linear Blending (SA-LB).} (a) Top view and (b) Front view of the grasping sequence. The U-shaped grippers represent the task space poses of the end-effector, where the solid black gripper denotes the human input ($\mathbf{u}_h^j$) at time $t=0$, and the red gripper indicates the autonomous grasping target ($\mathbf{u}_a^j$), which is kept constant during the approach for visualization. The gray border outlines the table workspace, and the green dots represent the perceived DLO point set $\mathcal{R}_t$. To clearly illustrate the system's behavior, we fixed the human input trajectory as a direct linear path from $\mathbf{u}_h^j$ to $\mathbf{u}_a^j$. The resulting trajectory, illustrated by the gradient line (black to red) and transparent intermediate grippers, shows the commanded poses of $\mathbf{u}_{\text{LB}}^j$. The pink box highlights a critical failure zone: because the blending is context-agnostic, the trajectory directly intersects the ungrasped rope. This unintended physical contact risks displacing the DLO, rendering the computed target $\mathbf{u}_a^j$ invalid and leading to task failure.}
    \label{fig:sa_lb}
    \vspace{-1em}
\end{figure}

\begin{figure}[t]
    \centering
    \includegraphics[width=1\linewidth]{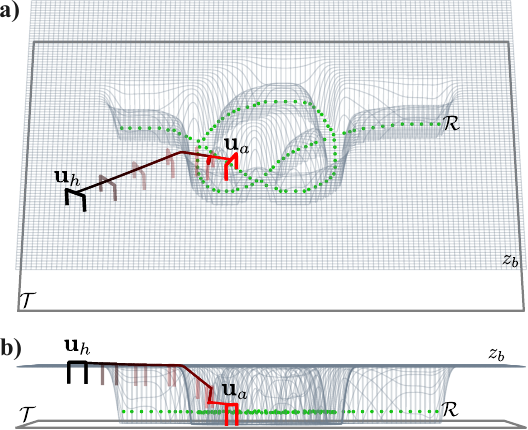}
    \vspace{-1em} \caption{\textbf{Shared Autonomy via Control Barrier Functions (SA-CBF).} (a) Top view and (b) Front view of the grasping sequence. Visual notations for the human input ($\mathbf{u}_h^j$), autonomous target ($\mathbf{u}_a^j$), and environment remain consistent with the \hyperref[fig:sa_lb]{Figure \ref*{fig:sa_lb}}. Subjected to the exact same fixed human input trajectory (a direct path from $\mathbf{u}_h^j$ to $\mathbf{u}_a^j$), the wireframe mesh visualizes the dynamic height-field safety barrier $z_b$, constructed using spatial funnels centered on the rope's perception points $\mathcal{R}_t$. The resulting trajectory demonstrates the corrective action of the SA-CBF formulation: rather than intersecting the rope, the commanded path $\mathbf{u}_{\text{CBF}}^j$ safely glides along the surface of the barrier boundary and down into the valid grasp funnel. The system successfully navigates the critical area without inadvertently sweeping or displacing the DLO, strictly preserving the spatial arrangement prior to the grasp.}
    \label{fig:sa_cbf}
    \vspace{-1em}
\end{figure}

At time $t$, the intent estimation module receives the DLO point set $\mathcal{R}_t$, the human input's current position $\mathbf{x}^{h,j}_t$, and the physical robot end-effector's current position $\mathbf{x}^{r,j}_t$ for each arm $j$. Our objective is to dynamically select a single intended target point $\hat{\mathbf{p}}^j_t \in \mathcal{R}_t$ for each arm that best matches the human intent while strictly preserving topological and temporal continuity.

Because bimanual DLO manipulation often occurs on flat work surfaces (e.g., tables), human operators naturally adopt a top-down grasping strategy to pick up the rope. To prevent the estimated target from jumping erratically across different parts of the rope due to sudden changes in the human trajectory or overlapping knot geometry, we implement a point selection algorithm equipped with a spatiotemporal hysteresis filter. The selector initially identifies the global nearest candidate in $\mathcal{R}_t$ to the human's target position $\mathbf{x}^{h,j}_t$ using the $L_1$~(Manhattan) distance. We use the $L_1$ distance rather than the standard Euclidean distance because it explicitly accounts for this top-down grasping behavior. By evaluating coordinate differences additively, the $L_1$ distance effectively decouples the horizontal planar alignment from the vertical approach depth, providing a target selection metric that naturally aligns with the decoupled Cartesian movements required for a top-down grasp.

Once an initial target is engaged, the algorithm enforces topological continuity based on the target selected in the previous time step, $\hat{\mathbf{p}}^j_{t-1}$. If no valid target is available from the previous frame (e.g., during initialization or when tracking is briefly lost), the system automatically falls back to using the globally nearest candidate. Otherwise, it constructs a local spatial neighborhood $\Omega_t$  (via KD-tree search) bounded by a maximum step radius around $\hat{\mathbf{p}}^j_{t-1}$. If the physical robot end-effector $\mathbf{x}^{r,j}_t$ is already close to this previous selection, the system restricts the new selection $\hat{\mathbf{p}}^j_t$ strictly to this local neighborhood. This localized hysteresis ensures that once the operator commits to a specific grasp region, the assistance remains stably anchored to that segment, avoiding discontinuous jumps between adjacent but topologically separate strands. A global target reselection based solely on proximity to $\mathbf{x}^{h,j}_t$ is only permitted when both the human input and the physical robot explicitly move away from the current region.

For every selected target point $\hat{\mathbf{p}}^j_t$, we continuously estimate the local rope tangent $\mathbf{t}_t$ by applying Principal Component Analysis (PCA) to its $k$-nearest neighbors in $\mathcal{R}_t$. Grasping thin objects directly from a flat table is notoriously challenging; human operators often hesitate out of fear of colliding the robot's gripper with the table surface. To alleviate this cognitive burden and ensure a safe grasp, we enforce a strict top-down orientation. We constrain the gripper's approach axis to remain perfectly vertical by setting the roll and pitch angles to zero relative to the task frame. The only actively controlled rotational degree of freedom is the yaw angle $\psi_t$, which is derived by projecting the local tangent $\mathbf{t}_t$ onto the horizontal plane so that the gripper fingers align perpendicularly to the DLO strand. This Euler angle configuration $(0, 0, \psi_t)$ is then converted into its corresponding quaternion $\mathbf{q}^{a,j}_t \in \mathbb{H}$. Consequently, the autonomous target pose $\mathbf{u}_a^j = (\mathbf{x}^{a,j}_t, \mathbf{q}^{a,j}_t)$ is fully defined for arm $j$, where the target position is the selected rope point ($\mathbf{x}^{a,j}_t = \hat{\mathbf{p}}^j_t$). By computing this independently for both arms, we formulate the complete bimanual autonomous target poses $\mathbf{u}_a = (\mathbf{u}_a^l, \mathbf{u}_a^r)$.

\begin{comment}Finally, to determine how aggressively the autonomous assistance should intervene, we compute an intent-based confidence score $\gamma_t \in [0, 1]$. This score scales inversely with the Manhattan distance between the human's input position and the selected rope point:
\[
\gamma_t=
\begin{cases}
0, & \|\hat{\mathbf{p}}_t - \mathbf{x}^h_t\|_1 > \delta_h,\\[2mm]
\max\!\left(0.1,\, 1-\dfrac{\|\hat{\mathbf{p}}_t - \mathbf{x}^h_t\|_1}{\delta_h}\right), & \text{otherwise},
\end{cases}
\]
where $\delta_h$ is the maximum assistance engagement radius (i.e., $\delta_h = 0.3$\,m). Consequently, when the user moves their controller close to the DLO, $\gamma_t$ approaches $1$, smoothly triggering the downstream framework to guide the gripper to the fully defined autonomous pose $u_a$. \end{comment}

\textbf{Control Barrier Function Based Shared Autonomy.}  \label{CBF}
To continuously synthesize the human operator's command $\mathbf{u}_h^j$ with the estimated autonomous target $\mathbf{u}_a^j$ for each arm $j \in \{l,r\}$, standard shared autonomy approaches often rely on Shared Autonomy via Linear Blending~(SA-LB), which we evaluate as a baseline in \hyperref[sec: exps]{Section~\ref*{sec: exps}}. SA-LB provides an attractor-based assistance baseline that pulls the end-effector toward an inferred target region. However, such methods lack the geometric awareness needed to prevent the robot's end-effector from inadvertently sweeping or displacing ungrasped DLO segments before a stable grasp is established, as indicated as a critical zone in \hyperref[fig:sa_lb]{Figure \ref*{fig:sa_lb}}.

To address this limitation, we propose a context-aware shared autonomy strategy based on Control Barrier Functions~(SA-CBF). In contrast to SA-LB, SA-CBF shapes motion through local geometric constraints, creating a geometry-aware funnel that supports precise approach and grasping without fully overriding the operator. A CBF provides mathematical guarantees that a system will not violate predefined physical boundaries by rendering a safe operational set forward invariant~\cite{ames2019cbf}. In our formulation, we use this property to project a virtual dynamic surface over the workspace, as shown in \hyperref[fig:sa_cbf]{Figure \ref*{fig:sa_cbf}}. This surface allows the operator to move freely in the open space above the table, but physically blocks the robot from descending—except through narrow, localized ``funnels'' explicitly centered over valid grasp targets. This intuitively enforces a precise, top-down grasping trajectory while rigorously preserving the spatial arrangement of the surrounding ungrasped rope strands during the approach.

To enforce this context-awareness, we construct a height-field barrier $z_b: \mathbb{R}^2 \to \mathbb{R}$ at each time step, which maps any horizontal coordinate $(x,y)$ to a minimum safe vertical clearance $z$. We initialize this virtual surface at a given nominal safe clearance height $z_0$ (which can also be estimated online via plane segmentation from the wrist-mounted RGB-D cameras). To safely permit top-down grasping, we introduce localized spatial funnels into this surface corresponding to the projected 2D coordinates $(p_{i,x}, p_{i,y})$ of every valid rope point $\mathbf{p}_i \in \mathcal{R}_t$. This creates a smooth lower geometric boundary that safely halts downward motion across the general table surface but allows deeper penetration exactly where the rope is located, as shown in \hyperref[fig:sa_cbf]{Figure \ref*{fig:sa_cbf}}.  The depth of the funnel at any given coordinate is determined by its proximity to the nearest valid rope point,  formulated as
\begin{equation*}
    z_b(x,y) = z_0 - \zeta \exp \left( -\frac{\min_{\mathbf{p}_i \in \mathcal{R}_t} \|(x,y) - (p_{i,x}, p_{i,y})\|_2^2}{2\sigma^2} \right),
\end{equation*}
where $\zeta$ is the maximum allowable penetration depth required for the gripper to reach the physical rope on the table (i.e., 0.02m), and $\sigma$ dictates the radial width of these approach funnels (i.e., 0.02). Let the physical robot end-effector's current position be $\mathbf{x}^{r,j}_t = [x^r, y^r, z^r]^\top$. The safety barrier function $h(\mathbf{x}^{r,j}_t)$ is defined as the signed vertical distance above this dynamic surface, given by
\begin{equation*}
    h(\mathbf{x}^{r,j}_t) = z^r - z_b(x^r, y^r) - \varepsilon,
\end{equation*}
where $\varepsilon$ is a strict minimal safety margin (i.e., 0.02). The safety set is rendered forward invariant as long as $h(\mathbf{x}^{r,j}_t) \ge 0$.

The entire action assistance framework is activated only when the human's input enters an active engagement zone, defined by the Euclidean distance $d^j = \|\mathbf{x}^{h,j}_t - \mathbf{x}^{a,j}_t\|_2 < \epsilon_{e}$,~(i.e.~$\epsilon_{e} = 0.3$m). Inside this region, we compute a nominal desired Cartesian velocity $\mathbf{v}_{\text{des}}$ that merges the human's input velocity $\mathbf{v}_{\text{human}}$ and an attractive velocity toward the autonomous target $\mathbf{v}_{\text{target}} = (\mathbf{x}^{a,j}_t - \mathbf{x}^{r,j}_t)/\Delta t$. To ensure the human retains the authority to break away from an assisted grasp, we evaluate the velocity alignments $\mathbf{v}_{\text{human}}^{\intercal} \mathbf{v}_{\text{target}}$. If the operator actively pulls away (i.e., the dot product is negative) and their movement is significant ($\|\mathbf{v}_{\text{human}}\| > 5$\,mm/s), we amplify their input by defining $\mathbf{v}_{\text{des}} = 2 \mathbf{v}_{\text{human}} + \mathbf{v}_{\text{target}}$. Otherwise, if the user stops or moves toward the target, the system prioritizes the approach by setting $\mathbf{v}_{\text{des}} = \mathbf{v}_{\text{human}} + \mathbf{v}_{\text{target}}$. We then solve for the optimal safe control velocity $\mathbf{v}^\star$ that tracks $\mathbf{v}_{\text{des}}$ as closely as possible while strictly satisfying the CBF constraint via a Quadratic Program defined~as
\begin{align*}
    \mathbf{v}^\star = \mathop{\arg\min}_{\mathbf{v} \in \mathbb{R}^3} \quad & \|\mathbf{v} - \mathbf{v}_{\text{des}}\|^2 \\
    \text{s.t.} \quad & \nabla h(\mathbf{x}^{r,j}_t)^\intercal \mathbf{v} + \kappa\big(h(\mathbf{x}^{r,j}_t)\big) \geq 0, \nonumber \\
                      & \|\mathbf{v}\|_\infty \le \mathbf{v}_{\max}, \nonumber
\end{align*}
where $\kappa(h) = \gamma h + \beta h^3$ is an extended class-$\mathcal{K}$ function with empirically tuned parameters (e.g., $\gamma = 100$, $\beta = 20$). The infinity-norm constraint $\|\mathbf{v}\|_\infty \le \mathbf{v}_{\max}$ caps the maximum velocity of any single Cartesian axis (i.e., $\mathbf{v}_{\max} = 0.2$\,m/s) to ensure safe execution. The inequality constraint actively modifies the command only when the robot attempts to penetrate the barrier outside of the valid rope regions, physically acting as a virtual force field that gently slides the gripper down the funnel toward the target rather than faulting. The optimized velocity yields the safe Cartesian target position $\mathbf{x}_{\text{CBF}}^j = \mathbf{x}_{t}^{r,j} + \mathbf{v}^\star \Delta t$.
Rather than selecting the task goal autonomously, SA-CBF intervenes only locally by guiding motion along geometry-consistent pathways once an intended interaction region has been inferred.

Because the height-field CBF strictly constrains 3D positional geometry, the gripper's orientation must be handled synchronously within the engagement zone. We compute a linear blending factor $\alpha(d^j) = \max(0, 1 - d^j/\epsilon_e)$. To ensure smooth execution, this factor is scaled by an interpolation weight (e.g., $0.1$). The assisted input orientation is then computed via Spherical Linear Interpolation (SLERP) between the previously commanded orientation $\mathbf{q}^{j}_{t-1}$ and the autonomous top-down target orientation $\mathbf{q}^{a,j}_t$, weighted by the smoothed blending factor, yielding $\mathbf{q}_{\text{CBF}}^j = \mathrm{SLERP}(\mathbf{q}_{t-1}^j, \mathbf{q}_t^{a,j};\alpha(d^j))$.

Finally, we address the post-grasp transition. During the assisted approach, the robot end-effector securely follows the CBF-constrained command, while the human operator's input remains safely floating above the table. If unconstrained control were instantly returned to the operator upon grasping the rope, the robot would violently snap back up to the human's current raw input pose $\mathbf{u}_h^j$. To prevent this spatial drift from causing a physical jerk, we initiate a handover phase when the gripper formally closes on the DLO. We interpolate from the rigidly constrained autonomous safe pose (recorded at the exact moment of grasp, denoted as $\mathbf{x}_{\text{grasp}}^j$ and $\mathbf{q}_{\text{grasp}}^j$) back to the raw human target over a brief temporal horizon:
\begin{align*}
    \mathbf{x}_{\text{CBF}}^j &= (1-\eta)\mathbf{x}_{\text{grasp}}^j + \eta \mathbf{x}^{h,j}_t, \\
    \mathbf{q}_{\text{CBF}}^j &= \mathrm{SLERP}(\mathbf{q}_{\text{grasp}}^j, \mathbf{q}^{h,j}_t; \eta),
\end{align*}
where $\eta$ ramps linearly from $0$ to $1$. This temporal blending ensures a seamless transition, enabling the user to perform free-form knot manipulation without autonomous intervention. Ultimately, whether actively assisting the approach, executing the handover, or granting unconstrained tracking post-grasp, the computed position and orientation continuously define the final commanded tracking pose $\mathbf{u}_\text{CBF}^j = (\mathbf{x}_\text{CBF}^j, \mathbf{q}_\text{CBF}^j)$. This pose is passed directly to the bimanual IK solver.

\section{Experiments} \label{sec: exps}
We designed a user study to validate the proposed framework and evaluate the efficacy of the assistive methods. 
Our experiments aim to answer the following core research questions:

\noindent\textbf{RQ1:} Does assistance improve DLO teleoperation performance relative to pure teleoperation?

\noindent\textbf{RQ2:} Does the effect of assistance depend on operator expertise?

\noindent\textbf{RQ3:} Does the effect of assistance depend on DLO properties such as stiffness and length?

\noindent\textbf{RQ4:} How do users perceive visual versus action assistance?

To address these questions, our experimental design maps three independent factors (assistance mode, human operator expertise group, and rope type) to objective metrics (success rate, task completion time) and subjective metrics (workload, helpfulness, preference).

\subsection{Experimental Setup}
\begin{figure*}[t]
\centering
    \includegraphics[width=1\linewidth]{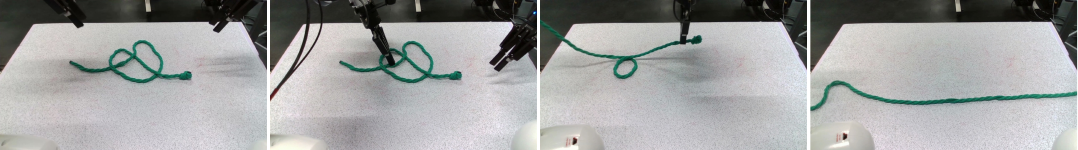}
    \vspace{-1em}
\caption{A sequential four-frame progression of an overhand knot untangling task. The manipulation is executed via our bimanual teleoperation setup and captured from the perspective of the fixed central workspace camera ($C_m$). Multiple demonstrations of the task are provided in the supplementary material.}
    \label{fig:sequence}
    
\end{figure*}

The bimanual teleoperation setup consists of two 7-DoF Franka Emika Panda robot arms equipped with Robotiq 2F-140 grippers, as shown in \hyperref[fig:main]{Figure \ref*{fig:main}}. Each robot wrist is equipped with a ZED X stereo camera for DLO State Estimation~($C_l, C_r$). A third fixed camera~($C_m$), Intel RealSense D435i, is positioned between the robots to provide a global view of the workspace for the human operator.

The teleoperation interface is built on the Unity Game Engine using the HTC Vive Pro VR system. The operator uses handheld VR controllers to provide 6-DoF input ($\mathbf{u}_h \in SE(3)$) and independently control the grippers on the left and right arms. A virtual plane in the VR environment displays the real-time RGB stream from the fixed central camera, ensuring the user has constant visual situational awareness, as shown in the operator view in \hyperref[fig:main]{Figure \ref*{fig:main}}.

The computational architecture is distributed: the vision pipeline runs on an NVIDIA RTX 4070Ti, providing continuous DLO state estimation at $\approx10$Hz. The robot kinematic controllers operate at $\approx100$Hz, while the VR framework runs at approximately $\approx90$Hz. Communication between the vision/control modules (ROS 1 Noetic) and the Unity interface is managed via the ROS-TCP-Connector and Google Remote Procedure Calls~(gRPC) for high-bandwidth visualization.

\subsection{Task and Procedure}
We selected a \textit{knot untangling} task as the evaluation benchmark, as it requires multi-step planning, bimanual coordination, and precise depth perception, as depicted in \hyperref[fig:sequence]{Figure~\ref*{fig:sequence}}. The specific goal is to untangle a loose overhand knot until the rope is free of self-intersections. The rope does not need to be perfectly straight; the success criterion is solely topological~(i.e., zero intersections).

We recruited 22 participants~(18 male, 4 female) with an average age of $39 \pm 8.61$ years (range: 24--56 years). Based on prior VR teleoperation experience, the pool was divided into 12 \textit{naive users}~(with limited or no experience) and 10 \textit{expert users}. Prior to the experiment, all participants were fully briefed on the study objectives and provided written informed consent. Participation was strictly voluntary, and individuals were explicitly instructed that they could take breaks or withdraw from the experiment at any time without consequence or the need to provide a reason. All data were collected and analyzed anonymously, ensuring no personally identifiable information was retained. The study protocol was approved by the Ethics Commission at TU Darmstadt. The experimental procedure began with a set of unrecorded training trials to familiarize participants with the system dynamics and VR interface. Following this training phase, each participant completed a total of 16 evaluation trials~(4 Control Conditions $\times$ 4 Rope Types). To mitigate learning effects and physical fatigue, which are highly significant factors in high-cognitive-load teleoperation tasks, the presentation order of the trials was systematically varied and randomized across participants. Each trial was constrained by a strict 3-minute time limit. A trial was formally marked as a failure if the participant exceeded this time limit or if a hardware ``safety stop'' was triggered three times during a single trial. Safety stops were automatically triggered by the robot's internal joint torque limits to prevent hardware damage. Common causes of these torque violations included the operator pulling the grasped rope in opposing directions with excessive force or inadvertently pressing the robot's end-effector against the table surface due to depth misestimation in unassisted modes.

\subsection{Study Conditions}

\begin{figure}[t]
    \centering
    \includegraphics[width=1\linewidth]{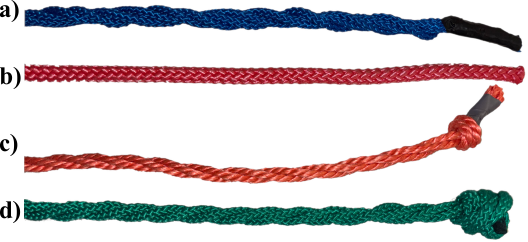}
    \vspace{-1em}
    \caption{The set of four DLOs used in the experimental validation includes (a) Blue rope, (b) Red rope, (c) Orange rope, and (d) Green rope. This figure shows the texture of the ropes, their colors, and their terminal features (e.g., electrical tape (Blue and Orange), heat-sealed tips (Red), knot (Green)).}
    \label{fig:Experimentropes}
    
\end{figure}

To evaluate the efficacy of the proposed framework, we conducted a user study comparing four distinct control modes. To ensure a fair baseline comparison and guarantee hardware safety, all four modes continuously pass their final commanded pose $u_f^j$ to the same underlying task-space inverse kinematics (IK) solver. This shared IK solver includes standard null-space projections for self-collision avoidance and a static geometric model of the workspace table to prevent environmental collisions. Consequently, while all modes share this fundamental low-level safety mechanism, they differ in how they assist the operator's high-level spatial perception and task-aware manipulation intent.

\noindent \textbf{Pure Teleoperation (PT):} As the unassisted baseline condition ($\mathbf{u}_f^j = \mathbf{u}_h^j$), the user relies solely on a compressed 2D video feed from the remote central camera ($C_m$) displayed within the VR headset. No depth augmentation or spatial guidance is provided.

\noindent \textbf{Visual Assistance (VA):} In addition to the standard 2D camera feed, the system dynamically renders the regularized DLO point set $\mathcal{R}_t$ directly into the user's 3D VR workspace. As the user moves their head, this spatial overlay provides critical motion parallax, yielding intuitive depth cues without any active robot intervention ($\mathbf{u}_f^j = \mathbf{u}_h^j$).

\noindent \textbf{Shared Autonomy via Linear Blending (SA-LB):} To assess the efficacy of our context-aware SA-CBF mode, we employ a baseline linear blending strategy. This mode inherits the 3D visual cues of VA and introduces active assistance by rendering a virtual ``ghost'' marker in the VR environment at the autonomous target pose $\mathbf{u}_a^j$ derived using our human intent estimation module (see \hyperref[sec:hie]{Section~\ref*{sec:hie}}) for each arm $j \in \{l, r\}$. To guide the physical robot toward this marker, we compute a scalar blending factor $\alpha(d^j) \in [0,1]$ derived from the Euclidean distance $d^j = \|\mathbf{x}^{h,j}_t - \mathbf{x}^{a,j}_t\|_2$ between the human and autonomous targets using a sigmoid function defined as
\begin{equation*}
    \alpha(d^j) = \frac{1}{1+\exp\!\big(-c(d^j/h-r)\big)},
\end{equation*}
where $h=0.6$ acts as a spatial scaling factor defining the maximum interaction range, $c=10$ dictates the steepness of the transition, and $r=0.4$ offsets the transition center. We blend the translation and orientation components separately to construct the final tracked pose $\mathbf{u}_{\text{LB}}^j = (\mathbf{x}_{\text{LB}}^j, \mathbf{q}_{\text{LB}}^j)$ using $\mathbf{x}_{\text{LB}}^j = \alpha(d^j) \mathbf{x}^{h,j}_t + (1-\alpha(d^j))\mathbf{x}^{a,j}_t$ and $\mathbf{q}_{\text{LB}}^j = \mathrm{SLERP}(\mathbf{q}^{a,j}_t, \mathbf{q}^{h,j}_t; \alpha(d^j))$. Consequently, when the human target is far from the intended grasp ($d^j \gg h \cdot r$), $\alpha \approx 1$ and control remains predominantly human-driven. As the user moves close to the target ($d^j \to 0$), $\alpha \approx 0$, and the autonomous assistance aligns the gripper towards the valid grasp target ($\mathbf{u}_f^j = \mathbf{u}_\text{LB}^j$).

\noindent \textbf{Shared Autonomy via Control Barrier Functions (SA-CBF):} This represents our proposed method. It uses the exact same 3D visual cues and the virtual ``ghost'' marker at $\mathbf{u}_a^j$ as the SA-LB condition, ensuring a consistent user interface. However, it replaces the context-blind linear blending with our geometry-aware SA-CBF formulation. This strategy guides the user toward the valid grasp target while actively ensuring the physical robot does not inadvertently displace ungrasped rope segments during the top-down approach phase ($\mathbf{u}_f^j = \mathbf{u}_\text{CBF}^j$).

\subsection{DLO Specifications} \label{subsec:dlo_specs}

\begin{figure}[t]
    \centering
    \includegraphics[width=1\linewidth]{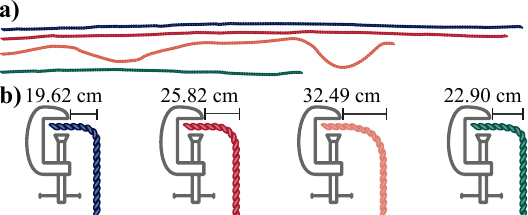}
    \vspace{-1em}
    \caption{To allow for a direct comparison of physical properties (length and bending behavior), (a) displays all four ropes stacked one on top of the other in a relaxed position (stretched tightly at both ends and laid on a flat surface) and (b) explains our measurements for DLO bending behavior using a large-deflection cantilever test. While the figure displays the measured horizontal projection distances ($b_\text{proj}$) for each rope, these values are mapped to the Heavy Elastica boundary value problem to estimate the effective flexural rigidities.}
    \label{fig:Experimentropescaliper}
    
\end{figure}

To evaluate robustness against physical variations, we used four custom-prepared ropes with distinct mechanical and visual properties, as depicted in \hyperref[fig:Experimentropes]{Figure \ref*{fig:Experimentropes}}. Each rope poses unique challenges regarding perception (homogeneity, length) and control (bending behavior). 
Flexural rigidity $EI$ is a key property of DLO manipulation because it governs resistance to bending during manipulation. To quantify this property, we adapted the standardization methodology of Garcia-Camacho et al.~\cite{garcia-camachoStandardizationClothObjects2024} from a 2D cloth drape test to a 1D large-deflection cantilever test. Each rope was clamped at one end, and the clamp contact length was subtracted from the total rope length to obtain the free-hanging length $L_\text{free}$. We then measured the horizontal projection distance $b_\text{proj}$ of the free end, as illustrated in \hyperref[fig:Experimentropescaliper]{Figure \ref*{fig:Experimentropescaliper}}.

Because highly flexible DLOs exhibit large downward deflections under their own weight, small-deflection linear beam theory is not applicable. Instead, we estimate the flexural rigidity ($EI$) by numerically solving the heavy-elastica differential equation
\begin{equation*}
    \frac{d^2\theta}{ds^2} + K(1-s)\cos\theta(s) = 0
\end{equation*}
subject to the clamped-free boundary conditions $\theta(0)=0$ and $\frac{d\theta}{ds}(1)=0$, where $s \in [0,1]$ is the normalized arc length along the rope, $\theta(s)$ is the local tangent angle, and
\begin{equation*}
    K = \frac{\lambda g L_\text{free}^3}{EI}
\end{equation*}
is a dimensionless gravity-loading parameter~\cite{Bickley01031934}. Here, $\lambda = m/L$ denotes the uniform linear mass density of the rope, where $m$ is the total rope mass and $L$ is the total rope length. In practice, we precompute the heavy-elastica response over a logarithmically spaced grid of $K$ values, yielding a lookup table for the normalized projection $\int_0^1 \cos\theta(s)\,ds$. The measured ratio $b_\text{proj}/L_\text{free}$ is then interpolated within this lookup table to estimate $K$, from which $EI$ is recovered numerically.

Using this methodology, we measured the linear mass density ($\lambda$) and flexural rigidity ($EI$) for our four experimental ropes. The \textbf{Blue rope} ($L=3.91$\,m, $m=140$\,g, diameter=1.27\,cm, $\lambda=0.0358$\,kg/m, $EI=0.0235$\,$\text{N}\text{m}^2$) is the longest in our set, featuring a three-strand braided structure and a tip wrapped in black electrical tape. While constructed from the same material as the \textbf{Green rope} ($L=2.21$\,m, $m=85$\,g, diameter=1.21\,cm, $\lambda=0.0385$\,kg/m, $EI=0.0170$\,$\text{N}\text{m}^2$), the blue rope's slightly larger diameter yields a higher area moment of inertia, resulting in a slightly higher absolute rigidity. The green rope is the most compliant in our set, exhibiting the lowest estimated flexural rigidity, and includes a small knot and electrical tape at its terminal. The \textbf{Red rope} ($L=3.76$\,m, $m=138$\,g, diameter=1.14\,cm, $\lambda=0.0367$\,kg/m, $EI=0.0390$\,$\text{N}\text{m}^2$) consists of a single thick strand with moderately high rigidity and a heat-sealed tip that creates high friction when separating overlapping segments. Finally, the \textbf{Orange rope} ($L=2.81$\,m, $m=115$\,g, diameter=1.17\,cm, $\lambda=0.0409$\,kg/m, $EI=0.0465$\,$\text{N}\text{m}^2$) possesses the highest flexural rigidity in the set, characterized by a slippery surface texture and a terminal capped with a tight knot and gray tape.

\subsection{Evaluation Metrics}
To explicitly map our experimental factors to measurable outcomes, we analyzed both objective task execution data and subjective user feedback.

\noindent \textbf{Objective Metrics:} We recorded the \textit{Task Completion Time}~(TCT) and the overall \textit{Success Rate}~(SR). A trial was considered successful if the operator completed the manipulation task~(e.g., untangling the knot) within the allocated time limit without triggering any hardware safety stops.

\noindent \textbf{Subjective Metrics:} Following each trial, participants completed a questionnaire designed to assess perceived workload and system usability. We employed the standard NASA-TLX subscales~\cite{hart1988development} alongside custom 7-point Likert scale questions designed to evaluate perceived control, system helpfulness, and rope predictability. The specific survey items and their corresponding abbreviations used in the analysis are detailed in \hyperref[tab:metrics]{Table~\ref*{tab:metrics}}.

\begin{table}[t]
\centering
\caption{Subjective Metrics and Survey Questions}
\label{tab:metrics}
\renewcommand{\arraystretch}{1.2}
\begin{tabular}{@{}lp{6.5cm}@{}}
\toprule
\textbf{Abbr.} & \textbf{Question / Metric} \\ \midrule
\multicolumn{2}{@{}l}{\textit{NASA-TLX Subscales}} \\
MD & How mentally demanding was the task? \\
PD & How physically demanding was the task? \\
TD & How hurried or time-pressured was the task? \\
PS & How successful were you in accomplishing the task goals? \\
EF & How hard did you have to work to accomplish the task? \\ \midrule
\multicolumn{2}{@{}l}{\textit{Custom Likert Items (7-Point Scale)}} \\
CTRL & I felt in control of the robot's behavior during this trial. \\
HELP & The system helped me accomplish the task in this trial. \\
UND & I understood what the system was doing and how it responded to my intent. \\
INT & The system interfered with or resisted my actions. \\
WANT & I would want this level of assistance again in similar tasks. \\
FGHT & I felt I had to fight against the system. \\
SAVE & The system saved me from making a mistake. \\ 
R-EXP & The rope behaved as I expected during this trial. \\
R-PROP & The rope's physical properties (e.g., length, stiffness) made the task harder.
\\
\bottomrule
\end{tabular}
\vspace{-1em}
\end{table}
\section{Results}
\label{sec:results}

Across all participants, assistance improved performance relative to pure teleoperation, but the relative benefit of each assistance mode depended strongly on both operator expertise and rope properties.

Because each participant completed multiple trials under varying conditions, we employed statistical models that accounted for within-subject dependence. Binary outcomes~(i.e., trial success or failure) were analyzed using Generalized Estimating Equations with a binomial distribution, a logit link function, and an exchangeable working correlation structure to estimate population-level effects. Continuous objective metrics, specifically TCT, were analyzed using linear mixed-effects models (LMMs). Subjective questionnaire data comprised NASA-TLX workload scores~(0--10 scale) and custom usability responses (1--7 Likert scale). Although these response formats are bounded and discrete, we analyzed them using LMMs. Treating Likert-type data as continuous variables is a widely accepted and robust practice in Human-Robot Interaction~(HRI) and behavioral sciences for capturing magnitude differences across experimental conditions~\cite{norman2010likert}.

In all mixed models, \textit{Participant ID} was included as a random intercept to account for individual baseline variances. Fixed effects included the \textit{Control Mode}~(PT, VA, SA-LB, SA-CBF) and \textit{Rope Type}~(Green, Blue, Red, Orange). Post-hoc pairwise comparisons between control modes were corrected using the Holm-Bonferroni method within each metric family to strictly control the family-wise error rate. Statistical significance thresholds were defined as $p < 0.05$ (*), $p < 0.01$ (**), and $p < 0.001$ (***). 

We organize the results around four research questions: overall efficacy of assistance (RQ1), expertise-dependent effects (RQ2), rope-dependent effects (RQ3), and subjective ratings and user preference (RQ4) as defined in \hyperref[sec: exps]{Section~\ref*{sec: exps}}.

\subsection{Overall Effect of Assistance (RQ1)}
To address \textbf{RQ1}, we first evaluate the overall effect of assistance on objective task performance, considering both TCT and SR across all participants.

\noindent \textbf{Success Rate:} The overall average success rates across all experimental conditions are detailed in \hyperref[tab:success_rate_all]{Table~\ref*{tab:success_rate_all}}. SA-CBF achieved the highest overall success rate (85.23\%). However, this improvement over the PT baseline (76.14\%) was not statistically significant at the population level ($p=0.30$). Notably, the common shared autonomy approach (SA-LB) achieved a slightly lower overall success rate (72.73\%) than unassisted PT. As we discuss further in \hyperref[sec:discussion]{Section~\ref*{sec:discussion}}, this modest performance drop under SA-LB substantiates our motivation for introducing SA-CBF. Thus, while assistance did not yield a statistically significant overall gain in success rate across the full cohort, the descriptive trend clearly highlights the benefit of context-aware action assistance over simple linear blending.

\begin{table}[t]
\centering
\caption{Overall Success Rate by Controller (All Users)}
\label{tab:success_rate_all}
\resizebox{\columnwidth}{!}{
\begin{tabular}{lcc}
\toprule
\textbf{Controller} & \textbf{Successful / Total Trials} & \textbf{Rate (\%)} \\ \midrule
Pure Teleoperation (PT) & 67/88 & 76.14 \\
Visual Assistance (VA)  & 66/88 & 75.00 \\
SA-LB                   & 64/88 & 72.73 \\
\textbf{SA-CBF}                 & \textbf{75/88} & \textbf{85.23} \\ \bottomrule
\end{tabular}
}
\vspace{-1em}

\end{table}
\begin{table}[t]
\centering
\caption{Task Completion Time (s) for Successful Trials (All Users)}
\label{tab:duration_results}
\resizebox{\columnwidth}{!}{
\begin{tabular}{lccc}
\toprule
\textbf{Controller} & \textbf{N} & \textbf{Mean $\pm$ Std} & \textbf{Median} \\ \midrule
Pure Teleoperation (PT) & 67 & $115.01 \pm 36.55$ & 116.0 \\
\textbf{Visual Assistance (VA)} & \textbf{66} & \textbf{90.65 $\pm$ 36.14} & \textbf{84.5} \\
SA-LB                   & 64 & $94.56 \pm 43.37$  & 95.5  \\
SA-CBF                  & 75 & $95.81 \pm 41.63$  & 91.0  \\ \bottomrule
\end{tabular}
}
\vspace{-1em}
\end{table}
\noindent\textbf{Task Completion Time:} TCT was analyzed exclusively for successful trials, with descriptive statistics reported in \hyperref[tab:duration_results]{Table~\ref*{tab:duration_results}}. Across the full population, all assistance modes significantly improved temporal efficiency relative to PT. The mean task duration for the unassisted PT condition was 115.01\,s. All assistive methods significantly reduced this time: VA to 90.65\,s~($p=0.0004^{***}$), SA-LB to 94.56\,s~($p=0.0068^{**}$), and SA-CBF to 95.81\,s~($p=0.0043^{**}$). No statistically significant differences were observed among VA, SA-LB, and SA-CBF when averaged across the full cohort. Thus, the primary overall impact of assistance was enhanced efficiency rather than a clear distinction between assistive modes.

\subsection{Expertise-Dependent Effects (RQ2)}
To address \textbf{RQ2}, a subgroup analysis reveals that SA-CBF produced the strongest gains for naive users, whereas VA remained highly competitive overall and was preferred by expert operators.

\begin{figure}[t]
    \centering
    \includegraphics[width=\linewidth]{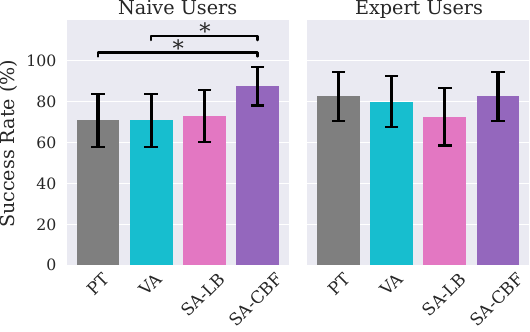}
    \vspace{-1em}
    \caption{Task success rates categorized by user expertise (Left: Naive, Right: Expert). Bar heights represent the mean success rate across trials, and error bars indicate the 95\% confidence intervals.}
    \label{fig:success_rate}
\end{figure}

\begin{figure}[t]
    \centering
    \includegraphics[width=\linewidth]{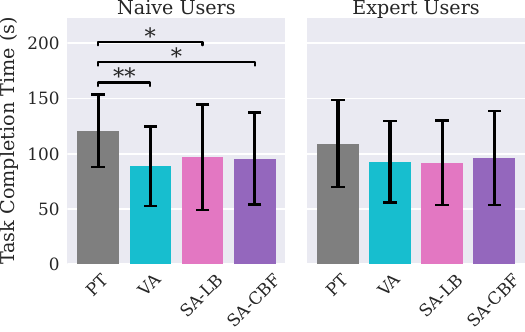}
    \vspace{-1em}
    \caption{Task Completion Times categorized by user expertise (Left: Naive, Right: Expert), computed only over successful trials. Bar heights represent the mean TCT across trials, and error bars indicate $\pm1$ standard deviation.}
    \label{fig:tct}
\end{figure}

For naive users~($N=12$), SA-CBF significantly improved the success rate to 87.50\%, compared to 70.83\% under PT ($p=0.0290^*$) and 66.67\% under VA~($p=0.0329^*$), as illustrated in \hyperref[fig:success_rate]{Figure~\ref*{fig:success_rate}}.  This advantage was also reflected in task completion time, as shown in \hyperref[fig:tct]{Figure~\ref*{fig:tct}}. Among successful trials, PT exhibited the highest mean completion time at 120.68\,s ($n=34$ successful trials). SA-CBF reduced this to 95.60\,s ($n=42$, $p=0.0176^*$), while VA reduced it to 88.79\,s ($n=34$, $p=0.0025^{**}$). These results indicate that naive users benefit substantially from assistance, with SA-CBF yielding the clearest gain in success rate and both SA-CBF and VA improving temporal efficiency.

Conversely, expert users~($N=10$) maintained high performance~($>80\%$) across most conditions naturally, with the notable exception of SA-LB, which dropped to 72.5\%. Experts were already faster in the unassisted PT condition, averaging 109.18\,s ($n=33$). While assistive methods lowered their mean completion times~(e.g., SA-CBF reduced it to 96.09\,s ($n=33$)), these differences were not statistically significant for the expert group. As illustrated in \hyperref[fig:success_rate]{Figure~\ref*{fig:success_rate}}, experienced operators were able to maintain strong performance without requiring action assistance. Taken together, these results suggest that expertise strongly mediates the utility of assistance, with physical intervention offering the greatest benefit to less experienced users.

\subsection{Rope and Material-Dependent Effects (RQ3)}
To address \textbf{RQ3}, we investigated how DLO properties, particularly length and bending stiffness, influenced manipulation difficulty and the relative effectiveness of different control modes. The objective performance metrics, grouped by rope type across the entire population, are summarized in \hyperref[tab:rope_quant]{Table~\ref*{tab:rope_quant}}. Although no statistically significant difference in overall success rate was detected, trials with the longest Blue rope showed a tendency toward lower success rates (68.18\%) and significantly longer trial durations than those with the other ropes.

\begin{table}[t]
\centering
\caption{Performance by Rope Type (All Users)}
\label{tab:rope_quant}
\begin{tabular}{lcc}
\toprule
\textbf{Rope Type} & \textbf{Success Rate (\%)} & \textbf{Mean TCT (s)} \\ \midrule
Green              & 87.50 & 89.61 \\
Orange             & 79.55 & 95.50 \\
Red                & 73.86 & 95.85 \\
Blue               & 68.18 & 118.53 \\ \bottomrule
\end{tabular}
\vspace{-1em}
\end{table}

\subsubsection*{Qualitative Perception of DLO Properties}
\begin{figure}[t]
    \centering
    \includegraphics[width=\linewidth]{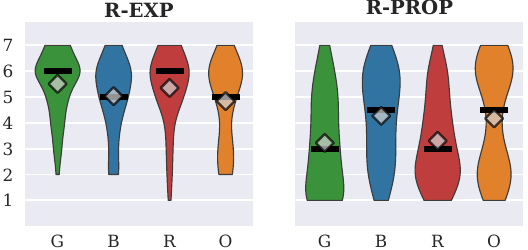}
    \vspace{-1em}
    \caption{Subjective rope metrics, as defined in \hyperref[tab:metrics]{Table~\ref*{tab:metrics}}. \textbf{Left (R-EXP):} Compared with the short, compliant Green rope, both the high-rigidity Orange rope and the much longer Blue rope were perceived as behaving less as expected during the trial. (Higher is better) \textbf{Right (R-PROP):} Participants reported that the physical properties of the Blue and Orange ropes made the task harder than those of the Green and Red ropes. (Lower is better)}
    \label{fig:likert_rope} 
\end{figure}

Subjective Likert ratings, shown in \hyperref[fig:likert_rope]{Figure~\ref*{fig:likert_rope}}, provide additional insight into how users perceived the handling characteristics of each rope. Across all participants, the Green rope received the highest R-EXP rating ($5.51$), indicating the closest agreement between observed rope behavior and user expectations. By contrast, the Orange rope received significantly lower R-EXP scores ($4.84$, $p=0.0078^{**}$ vs.\ Green), consistent with its substantially higher flexural rigidity. Expertise-specific analysis further showed that expert users rated both the Blue rope and the Orange rope as behaving significantly less than expected than the Green rope ($p=0.0347^{*}$ and $p=0.0370^{*}$, respectively). For the Orange rope, this reduction is consistent with its high rigidity. For the Blue rope, which differs from the Green rope by substantially greater length, the reduced R-EXP is likely driven by a combination of these factors rather than by a single physical property.

For R-PROP, both the Orange and Blue ropes were rated as showing significantly greater property-related task difficulty than the Green and Red ropes ($p < 0.01$), indicating a stronger contribution of their physical properties to perceived task difficulty. These effects are consistent with two distinct challenge profiles: high rigidity for the Orange rope and large-scale deformation for the Blue rope, due to its greater length and compliance relative to the stiffer Red rope. The bimodal distribution observed for the Orange rope in \hyperref[fig:likert_rope]{Figure~\ref*{fig:likert_rope}} becomes clearer when separating users by expertise and corresponds to the largest rating variability across both groups. Naive users more often associated the stiffest rope with increased difficulty, whereas expert users more often associated the longest rope with increased difficulty.

\noindent\textbf{Performance on Stiff DLOs (Red, Orange):} The clearest benefits of SA-CBF were observed for the stiffer objects. Due to its high stiffness (high $EI$), it naturally prevents rope segments from resting in close proximity or undergoing complex deformations. Consequently, the estimated DLO state and, by extension, the inferred human intention, become more accurate. This improved estimation enables the control barrier function in SA-CBF to construct more clearly defined, conservative safety boundaries that effectively guide user interaction while reducing conflicts or interference with neighboring segments.
For the \textbf{Red Rope} (high inter-segment friction, moderate stiffness), users struggled significantly to secure precise grasps under unassisted PT, resulting in a mean task duration of 122.07 s. SA-CBF drastically reduced this time to 83.63 s ($p=0.0005^{***}$). 
For the \textbf{Orange Rope} (highest stiffness), naive users struggled heavily with the PT and VA, achieving only a 66.7\% success rate. SA-CBF was highly effective in this condition, achieving a 100\% success rate for Naive users ($p < 0.0001^{***}$ vs. PT). This suggests that geometry-aware action assistance is especially useful when stiffness and tension make precise grasping difficult.

\begin{figure}[t]
    \centering
    \includegraphics[width=\linewidth]{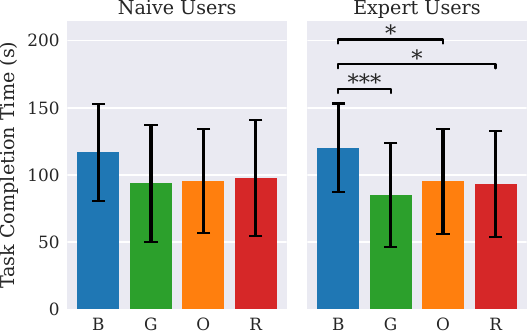}
    \vspace{-1em}
    \caption{Task Completion Time (TCT) by Rope Type. The longest DLO (Blue) required significantly more manipulation time than the shorter baseline (Green, $p=0.0001$), clearly demonstrating the severe impact of object length on teleoperation complexity. (Left: Naive, Right: Expert) Bar heights represent the mean TCT across trials, and error bars indicate $\pm1$ standard deviation.}
    \label{fig:duration_rope}
\end{figure}

\noindent\textbf{Performance on Compliant DLOs (Blue, Green):} For highly deformable objects, the benefits of action assistance show a more complex profile, and VA emerged as a particularly competitive strategy.
The \textbf{Blue Rope} (long) was the most chaotic DLO, requiring significantly more time (118.53 s) than the Green rope ($p=0.0001^{***}$, see \hyperref[fig:duration_rope]{Figure~\ref*{fig:duration_rope}}). In this condition, action assistance methods did not improve the success rate relative to the baseline. The long, compliant rope also exposed an important limitation of local perception. Untangling required wide, rapid motions, and the rope frequently moved out of the wrist-camera field of view. Without a consistent visual context, action assistance could only provide limited local guidance. Conversely, VA proved highly effective; for Naive users, VA significantly reduced the task duration to 99.4 s compared to the 142.0 s PT baseline ($p=0.0165^{*}$). For the \textbf{Green Rope}~(short, compliant baseline), a ceiling effect was observed. Success rates were uniformly high ($>83\%$). However, for Expert users, VA achieved a perfect 100\% success rate and the fastest TCT (71.4 s) in Green rope. Overall, these findings suggest that visual support is especially beneficial when the handled DLO is long and operated by expert users. \hyperref[tab:synthesis_matrix]{Table~\ref*{tab:synthesis_matrix}} summarizes assistance performance across operator expertise levels and DLO properties. Based on the observed trends, the most appropriate control mode is identified for each condition.

\begin{table*}[t] 
\centering 
\caption{Synthesis of Assistance Performance by Expertise and DLO Properties} \label{tab:synthesis_matrix} \renewcommand{\arraystretch}{1.3} \small 
\begin{tabularx}{\textwidth}{@{} l XXXX @{}} 
\toprule & \multicolumn{4}{c}{\textbf{DLO Properties and Rope Type}} \\
\cmidrule(lr){2-5}
\textbf{Operator Expertise} 
  & \textbf{Short/Compliant} (Green) 
  & \textbf{Long/Moderate-Compliant} (Blue) 
  & \textbf{Moderate-Long/Moderate-Stiffness} (Red) 
  & \textbf{Moderate-Short/High-Stiffness} (Orange) \\
\midrule
\textbf{Naive} & \textbf{SA-CBF} / \textbf{VA} \newline Optimal SR \& Speed & \textbf{VA} \newline 
Mitigates local sensing loss & \textbf{SA-CBF} \newline Aids target capture & \textbf{SA-CBF} \newline 100\% SR; overcomes tension \\ 
\addlinespace \textbf{Expert} & \textbf{VA} \newline Fastest; preserves authority & \textbf{VA} / \textbf{PT} \newline Manual global control & \textbf{SA-CBF} / \textbf{VA} \newline CBF speed vs. VA authority & \textbf{VA} \newline Fastest; preserves authority\\ 
\bottomrule 
\multicolumn{5}{@{} p{\textwidth} @{}}{\footnotesize \textit{Note:} This table synthesizes descriptive trends from success rates, task times, and subjective acceptance within each expertise subgroup. Cells listing multiple modes (e.g., `SA-CBF / VA') indicate that both showed favorable descriptive trends in that condition; this does not indicate statistical equivalence, as pairwise significance was not established for all such combinations.} \\ 
\end{tabularx} 
\vspace{-1em}
\end{table*}

\subsection{Subjective Ratings and Preference (RQ4)}

\begin{figure*}[t]
    \centering
    \includegraphics[width=1\textwidth]{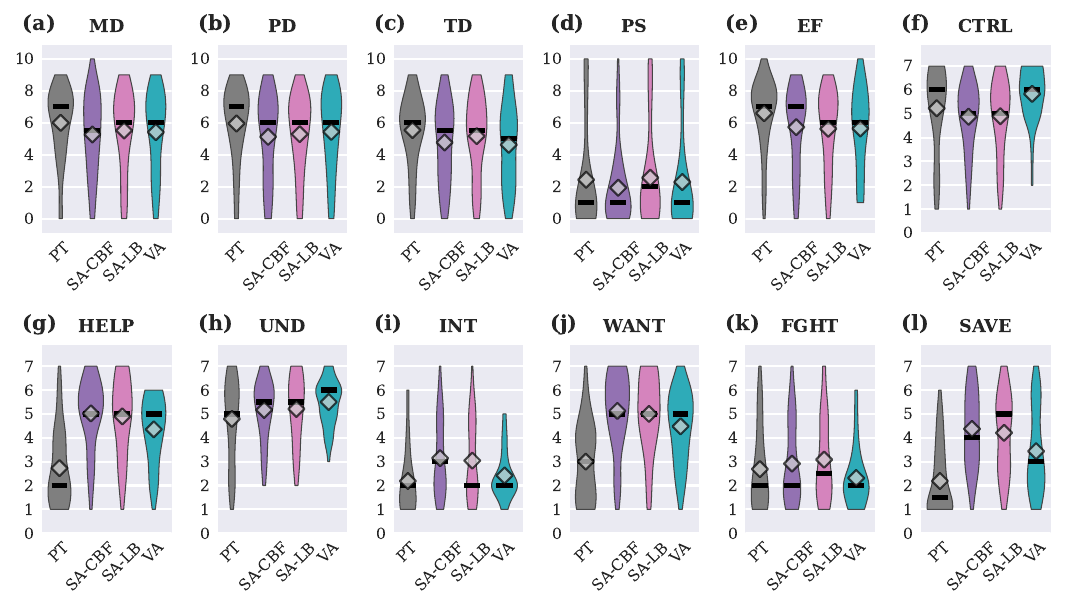}
    \vspace{-1em}
    \caption{Subjective evaluation distributions for \textbf{Naive Users} ($N=12$). The 2$\times$6 grid displays workload and authority metrics (MD, PD, TD, PS, EF, CTRL) in the top row, and system perception and preference metrics (HELP, UND, INT, WANT, FGHT, SAVE) in the bottom row. Each violin plot illustrates the data density, alongside the mean (light grey diamond) and median (thick black horizontal line). Naives exhibited high acceptance of physical assistance, rating SA-CBF highest in Helpfulness and safety (SAVE), with minimal perceived interference (INT) or conflict (FGHT) compared to the unassisted baseline.}
    \label{fig:subj_naive}
    
\end{figure*}

\begin{figure*}[t]
    \centering
    \includegraphics[width=1\textwidth]{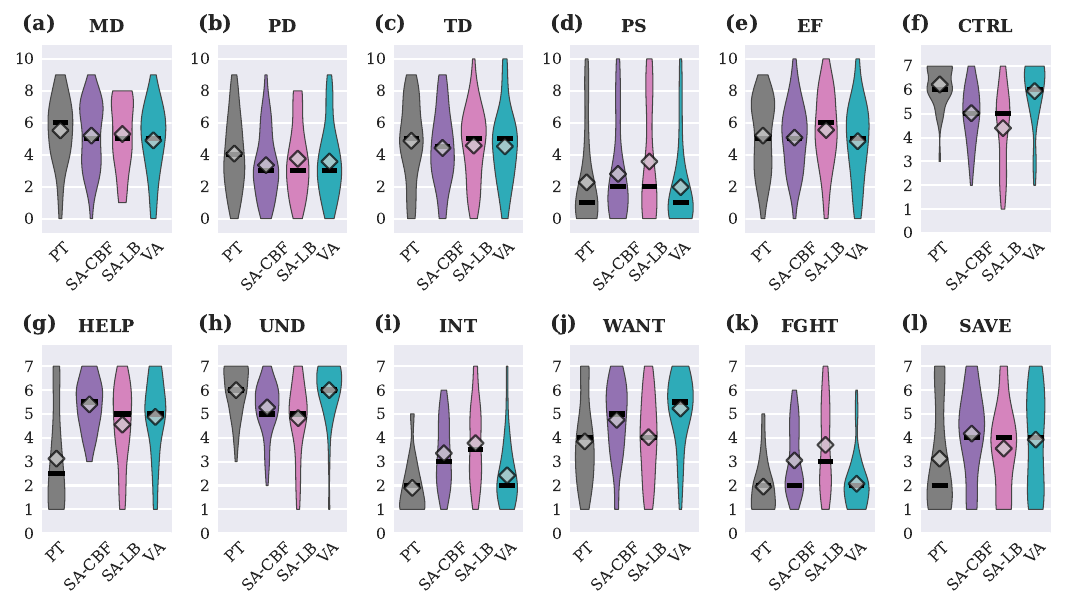}
    \vspace{-1em}
    \caption{Subjective evaluation distributions for \textbf{Expert Users} ($N=10$) across the same 12 metrics. As depicted in \hyperref[fig:subj_naive]{Figure~\ref*{fig:subj_naive}}, plots display data density, mean (light grey diamond), and median (thick black horizontal line). In contrast to naive users, experienced operators reported a significant loss of authority (CTRL) and an increase in system conflict (FGHT) when subjected to physical intervention (SA-CBF and SA-LB). Consequently, experts demonstrated a strong overall preference (WANT) for purely Visual Assistance (VA), which preserved their baseline proficiency while providing spatial understanding without physical interference.}
    \label{fig:subj_expert}
\end{figure*}

\begin{table*}[t]
\centering
\caption{Comprehensive Subjective Metrics for All Users ($N=22$)}
\label{tab:subj_results_all}
\renewcommand{\arraystretch}{1}
\resizebox{\linewidth}{!}{%
\begin{tabular}{lccccccccccccccc}
\toprule
 & \multicolumn{5}{c}{\textbf{NASA-TLX Workload (0--10)}} & & \multicolumn{5}{c}{\textbf{System Perception (1--7)}} & & \multicolumn{3}{c}{\textbf{Preference (1--7)}} \\ 
\cmidrule{2-6} \cmidrule{8-12} \cmidrule{14-15}
\textbf{Controller} & \textbf{MD} & \textbf{PD} & \textbf{TD} & \textbf{PS} & \textbf{EF} & & \textbf{CTRL} & \textbf{INT} & \textbf{FGHT} & \textbf{SAVE} & \textbf{UND} & & \textbf{HELP} & \textbf{WANT} \\ \midrule
Pure Teleoperation (PT) & 5.78 & 5.10 & 5.24 & 2.36 & 5.97 & & \textbf{5.68} & \textbf{2.06} & 2.35 & 2.61 & 5.34 & & 2.91 &  3.39 \\
Visual Assistance (VA)  & 5.18 & 4.59 & 4.58 & 2.15 & 5.28 & & \textbf{5.89} & 2.42 & \textbf{2.20} & 3.66 & \textbf{5.73} & & 4.59 &  \textbf{4.82} \\
SA-LB                   & 5.42 & 4.59 & 4.90 & 3.02 & 5.59 & & 4.67 & 3.38 & 3.36 & 3.91 &5.03 & & 4.74 & 4.56 \\
\textbf{SA-CBF}         & 5.24 & \textbf{4.32} & 4.61 & 2.33 & 5.43 & & 4.94 & 3.24 & 2.98 & \textbf{4.28} & 5.22 & & \textbf{5.19} &  \textbf{4.95} \\ 
\bottomrule
\multicolumn{16}{l}{\footnotesize \textit{Note:} Bold indicates the optimal condition(s) per metric. Multiple values are bolded if there is no statistical significance between top performers.} \\
\multicolumn{16}{l}{\footnotesize Workload \& Interference (MD, PD, TD, PS, EF, INT, FGHT): Lower is better. Utility \& Preference (CTRL, SAVE, UND, HELP, WANT): Higher is better.}
\end{tabular}%
}

\end{table*}
To address \textbf{RQ4}, we analyzed participants' subjective perception of the system using NASA-TLX workload indices and custom usability metrics, listed in \hyperref[tab:subj_results_all]{Table~\ref*{tab:subj_results_all}}. The distributions of these responses for the Naive and Expert groups are visualized in \hyperref[fig:subj_naive]{Figure~\ref*{fig:subj_naive}} and \hyperref[fig:subj_expert]{Figure~\ref*{fig:subj_expert}}.

\noindent \textbf{Workload and Effort:} Across the full population, the task was perceived as mentally demanding regardless of the controller, though SA-CBF recorded the lowest median mental demand~($MD=5.0$) compared to all other modes~($MD=6.0$). For the total population and the naive users, SA-CBF significantly reduced \textit{Physical Demand}~(PD) compared to the unassisted PT baseline~($p=0.0032^{**}$ and $p=0.0206^{*}$, respectively). Notably, expert users reported exerting significantly more effort~(EF) and feeling less successful~(PS) when using the context-agnostic SA-LB compared to VA~($p=0.0193^{*}$), suggesting that context-blind action assistance actively frustrates experienced operators.

\noindent \textbf{Control and Helpfulness:} A clear trade-off emerged between the sensation of authority and the utility of the system. Whereas the control modes that do not intervene in human input (PT, VA) inherently yield the highest perceived level of Control~(CTRL), experts reported a significantly greater perceived loss of authority under SA-CBF than under PT ($p=0.0001^{***}$). Interestingly, even though VA provides no physical intervention, users rated its Interference~(INT) significantly lower than PT, suggesting that increased visual clarity reduces the perceived cognitive overhead of the teleoperation mapping. In contrast, naive users were more inclined to follow the guidance. Across the full cohort, SA-CBF was rated as significantly more helpful (HELP) and better at preventing errors~(SAVE) than the PT baseline ($p < 0.001$). For experts, SA-CBF was significantly more helpful than SA-LB ($p=0.0333^{*}$), proving the value of context-awareness over blind attractors. This pattern was also reflected in the ``Fighting''~(FGHT) metric; users reported greater conflict with SA-LB, whereas SA-CBF reduced perceived system conflict relative to linear assistance.

\noindent\textbf{System Understandability:} The ``Understandability''~(UND) metric reveals a clear asymmetry between user groups. Across the full cohort, VA achieved the highest perceived understandability (mean = 5.73), significantly outperforming SA-LB (p = 0.0021**) and SA-CBF (p = 0.0418*). Naive users showed no significant differences across conditions, suggesting limited prior expectations made all modes similarly legible. Expert users, however, showed a pronounced drop in action assistance. While PT and VA both achieved a mean of 6.00, SA-LB fell to 4.82 (p = 0.0001***) and SA-CBF to 5.28 (p = 0.0272*). This mirrors the CTRL and FGHT findings — experts with well-formed teleoperation priors found physical interventions harder to anticipate. While SA-CBF showed a numerically higher mean than SA-LB among experts (5.28 vs. 4.82), this difference did not reach statistical significance.

\noindent \textbf{User Preference:} Preference was highly divided by expertise. Naive users strongly preferred SA-CBF, followed by SA-LB and VA. However, experts preferred VA, followed by SA-CBF, with SA-LB being the least desired assistive mode. Across all users, the WANT metric was significantly higher for assisted conditions than for PT ($p < 0.001$), indicating that some form of assistance was generally preferred over fully unassisted teleoperation. Taken together, the subjective results mirror the objective findings: naive users favor physically supportive action assistance, whereas experts prefer visual assistance that improves perception without reducing control authority.

\section{Discussion}
\label{sec:discussion}
Our results show that while shared autonomy can improve DLO manipulation, its effectiveness depends strongly on both the DLO's physical properties and the human operator's prior teleoperation experience. In this section, we interpret these findings through the geometry-aware shared autonomy, human-robot interaction, and physical DLO properties.

\noindent\textbf{The ``Funnel'' vs. The ``Magnet'' (SA-CBF vs. SA-LB):} A central finding of our study is the clear difference in both objective performance and user perception between the two action-assistance modes. Because both methods utilized the same multi-view vision pipeline and human intent estimation, the observed differences point to a fundamental distinction in how these control formulations modulate human input. Linear Blending (SA-LB) acts as a target-directed attractor that pulls the user toward an inferred goal region without explicitly accounting for local DLO geometry. In contrast, the SA-CBF formulation acts as a geometry-aware local constraint mechanism, functioning like a funnel anchored directly to the DLO, as described in \hyperref[CBF]{Section~\ref*{CBF}}. Because approaching the DLO from the side risks accidental collisions that could displace the rope, this funnel explicitly constrains unsafe movements.
Furthermore, by dynamically modulating the commanded velocity, SA-CBF adapts its assistance based on human input. When human input is minimal or aligned with the goal, the controller guides the end-effector toward the inferred target pose. Conversely, if the human intentionally applies a control input to navigate away from the target, the controller correspondingly amplifies this velocity to facilitate a rapid escape from the local region, all while continuously enforcing the forward invariance of the safe set. These results suggest that, in deformable environments, constraint-based guidance is more compatible with human operators than simple attractor-based assistance. For example, if a user intends to approach from a specific angle to avoid a complex knot topology, SA-LB actively resists this intent if it diverges from the shortest linear path, leading to high perceived interference (INT). 
More generally, this comparison explains why SA-LB slightly underperformed the pure teleoperation baseline for the full cohort; the form of assistance matters as much as its presence, and physically intervening systems must account for local geometry to avoid causing disruption.

\noindent\textbf{The Expertise Gap and Internal Priors:} Our subgroup analysis reveals a clear expertise-dependent pattern in teleoperation performance. Naive users, lacking a stable internal prior for sensitive teleoperation, benefited strongly from action assistance. For these users, SA-CBF served as a substantial performance equalizer, narrowing the gap with more experienced users and improving success rates. Conversely, expert operators derived less benefit from physical intervention. Interestingly, while experts explicitly acknowledged that SA-CBF was helpful and actively prevented errors (SAVE), as shown in \hyperref[fig:subj_expert]{Figure~\ref*{fig:subj_expert}}, they still strongly preferred Visual Assistance (VA). This preference aligns with recent shared autonomy literature, as experts already possess highly optimized internal spatial priors for task execution; they frequently reject physical intervention. For example, Belsare et al. similarly documented expert users who indicated a strict preference for pure teleoperation even when near-optimal assistance was available \cite{belsareZeroShotUserIntent2025}. When an autonomous agent's constraint, even when mathematically well motivated, alters an expert's planned trajectory or grasp approach, the expert perceives the intervention as a disturbance. Accordingly, our results suggest that action assistance is most valuable for bridging the novice skill gap, whereas expert users more often prefer \textit{vision assistance} (VA) over \textit{action assistance}. 
Since VA also uses the same multi-view DLO state estimation pipeline, these findings further indicate that the perception module is sufficiently informative to support online assistance across the tested manipulation conditions. This distinction between action support for naive users and perception support for expert users is one of the main design implications of the study.

\noindent\textbf{Hardware Limits and Object Compliance:} The interaction analysis highlights a critical intersection between physical object properties and hardware limitations. SA-CBF proved to be a stabilizer for stiff, difficult-to-grasp ropes~(Red and Orange), making precise grasping significantly faster and easier. However, for the very long and compliant Blue rope, visual assistance (VA) outperformed action assistance. This outcome likely reflects an interaction between the assistance's characteristics and the hardware's sensing limits. Because untangling long DLOs is time-consuming, operators frequently move the robot rapidly across the workspace. During these wide, sweeping motions, the long rope often fell outside the camera's limited field of view (FoV). In the absence of a complete visual context, the vision processing pipeline cannot reliably compute local barrier functions. These observations suggest that when hardware constraints, such as limited camera field of view, are routinely challenged by large-scale deformations, global visual cues can be more robust than localized physical constraints. In this sense, the results highlight an important boundary of the current system, which means action assistance is most effective when the geometry remains observable. In contrast, global visual support becomes increasingly advantageous as the object moves beyond the reliably observed workspace. This FoV constraint also highlights an important limitation of our evaluation of the perception pipeline. A rigorous quantitative comparison against existing DLO tracking methods was not feasible in our setting, as these methods assume fixed single-camera configurations and struggle with the high topological complexity and frequent self-intersections inherent to our task, conditions that are systematically violated in our bimanual, high-occlusion knot-untangling task. While the absence of dense ground-truth annotations prevents quantitative analysis (such as IoU), the reliability of our state estimation is indirectly supported by downstream task performance: the statistically significant decreases in task completion time achieved with VA and SA-CBF suggest that the pipeline yields geometric estimates that are sufficiently accurate within the observable workspace.
\section{Conclusion}
\label{sec:conclusion}

In this paper, we presented \textit{AssistDLO}, an assistive teleoperation framework for deformable linear object (DLO) manipulation that combines multi-view DLO state estimation, visual assistance, and a context-aware shared-autonomy controller based on Control Barrier Functions. Beyond the system itself, the main contribution of this work is the empirical finding that effective assistance for DLO teleoperation depends on both the operator and the object.

Through a comprehensive user study with 22 participants, we showed that different forms of assistance have distinct effects across expertise levels. In particular, conventional shared-autonomy approaches such as linear blending can degrade performance for expert users, who already possess strong teleoperation priors and often prefer visual feedback over physical intervention. In contrast, naive users consistently benefit from assistance, with our SA-CBF method providing the strongest improvements. By guiding motion through geometry-aware local constraints rather than simply attracting the robot toward a target, SA-CBF reduces perceived physical demand and improves performance while preserving operator authority.
Most importantly, it substantially narrows the expertise gap, increasing novice success from 70.83\% in pure teleoperation to 87.50\%.

Our results also show that DLO properties strongly influence the effectiveness of assistance. Long and highly compliant ropes are harder to manipulate, while shorter and stiffer ropes are more responsive to action assistance. In particular, SA-CBF is especially beneficial for stiffer ropes. In contrast, its effectiveness decreases for long, highly compliant objects due to increased unpredictability and the limited local context available from wrist-mounted cameras. These findings indicate that DLO teleoperation cannot be addressed with a single fixed assistance strategy. Instead, assistance should be adaptive to operator expertise, object mechanics, and perceptual conditions.

The multi-view DLO state estimate serves as the geometric basis for all assistance modes and, together with human intent estimation and point selection, enables online grasp support in cluttered and unconstrained manipulation settings. This support is particularly relevant in DLO tasks, where inaccurate grasping or unintended topology changes can quickly lead to task failure.

The presented approach also has several limitations. The perception scope is limited as the current pipeline targets online geometric estimation for assistance and assumes a single distinguishable DLO. The locality of action assistance is another limitation, as SA-CBF is designed for local grasp support rather than full task-level planning. Additionally, hardware and field-of-view constraints pose challenges, as long compliant ropes can leave the wrist-camera field of view during large workspace motions, reducing the reliability of local action guidance. Furthermore, the task scope is limited, as although we varied rope properties, our experiments focused on overhand-knot untangling and should be extended to other DLO tasks and to more controlled ablations of stiffness, length, friction, and mass. These limitations reflect the present scope of the system rather than fundamental obstacles to the proposed assistance paradigm.

These limitations suggest several promising directions for future research. A key next step is adaptive arbitration between assistance modes, where the system estimates operator proficiency, task phase, and DLO behavior in real time and continuously adjusts the balance between visual and action assistance.
More robust perception, including memory over temporary occlusions and improved handling of visually ambiguous objects, will also be important. In the longer term, integrating more robust DLO models and knot-aware planning may enable task-level assistance that not only supports grasping but also provides online guidance and feedback on manipulation progress.

Overall, our findings suggest that DLO teleoperation should not rely on a single fixed assistance strategy. Instead, effective assistance must be matched to the operator, the task phase, and the physical behavior of the manipulated object. We hope that \textit{AssistDLO} contributes to the development of future adaptive shared-autonomy systems that combine safety, usability, and task effectiveness in complex deformable-object manipulation.
\section*{Acknowledgments}
The author Berk Güler is funded by Honda Research Institute Europe GmbH. \section*{Conflict of Interest}
The authors declare no conflict of interest.
\section*{Data Availability Statement}
Data will be made available on request.
\section*{Ethics Approval Statement}
Appropriate Ethical Committee approval was obtained for the user study ($N=22$) presented in this work from the Ethics Committee of TU Darmstadt (Application No. EK 66/2023). All participants provided written informed consent prior to participation in accordance with the approved protocol.
\appendix
\section*{Supplementary Material}
A supplementary video demonstrating the AssistDLO framework and representative trials across all experimental conditions is provided as supplementary material.

% --- START OF END-MATTER ---
% --- START OF END-MATTER ---

% --- END OF END-MATTER ---

\bibliography{references}{}

@article{draganTeleoperationIntelligentCustomizable2013,
  title = {Teleoperation with {{Intelligent}} and {{Customizable Interfaces}}},
  author = {Dragan, Anca D and Siddhartha Srinivasa, Siddhartha and Kenton Lee, Kenton},
  year = {2013},
  month = jun,
  journal = {Journal of Human-Robot Interaction},
  volume = {2},
  number = {2},
  pages = {33--79},
  issn = {2163-0364},
  doi = {10.5898/JHRI.2.2.Dragan},
  urldate = {2024-02-01},
  langid = {english},
}

@article{fuchsGazeBasedIntentionEstimation2021,
  title = {Gaze-{{Based Intention Estimation}} for {{Shared Autonomy}} in {{Pick-and-Place Tasks}}},
  author = {Fuchs, Stefan and Belardinelli, Anna},
  year = {2021},
  month = apr,
  journal = {Frontiers in Neurorobotics},
  volume = {15},
  pages = {647930},
  issn = {1662-5218},
  doi = {10.3389/fnbot.2021.647930},
  urldate = {2024-02-01},
  langid = {english},
}

@article{gottardiSharedControlRobot2022,
  title = {Shared {{Control}} in {{Robot Teleoperation With Improved Potential Fields}}},
  author = {Gottardi, Alberto and Tortora, Stefano and Tosello, Elisa and Menegatti, Emanuele},
  year = {2022},
  month = jun,
  journal = {IEEE Transactions on Human-Machine Systems},
  volume = {52},
  number = {3},
  pages = {410--422},
  issn = {2168-2291, 2168-2305},
  doi = {10.1109/THMS.2022.3155716},
  urldate = {2024-02-01},
  langid = {english},
}

@article{hauserRecognitionPredictionPlanning2013,
  title = {Recognition, Prediction, and Planning for Assisted Teleoperation of Freeform Tasks},
  author = {Hauser, Kris},
  year = {2013},
  month = nov,
  journal = {Autonomous Robots},
  volume = {35},
  number = {4},
  pages = {241--254},
  issn = {0929-5593, 1573-7527},
  doi = {10.1007/s10514-013-9350-3},
  urldate = {2024-02-01},
  langid = {english},
}

@misc{javdaniSharedAutonomyHindsight2017,
  title = {Shared {{Autonomy}} via {{Hindsight Optimization}} for {{Teleoperation}} and {{Teaming}}},
  author = {Javdani, Shervin and Admoni, Henny and Pellegrinelli, Stefania and Srinivasa, Siddhartha S. and Bagnell, J. Andrew},
  year = {2017},
  month = may,
  number = {arXiv:1706.00155},
  eprint = {1706.00155},
  primaryclass = {cs},
  publisher = {arXiv},
  urldate = {2024-04-03},
  archiveprefix = {arxiv},
  langid = {english},
}

@inproceedings{manschitzSharedAutonomyIntuitive2022,
	title = {Shared Autonomy for Intuitive Teleoperation},
	author = {Simon Manschitz AND Dirk Ruiken},
	year = {2022},
	month = {May},
	publisher = {-},
	booktitle = {ICRA Workshop: Shared Autonomy in Physical Human-Robot Interaction: Adaptability and Trust}
}

@incollection{draganFormalizingAssistiveTeleoperation2013,
    author = {Dragan, Anca D. and Srinivasa, Siddhartha S.},
    isbn = {9780262315722},
    title = {Formalizing Assistive Teleoperation},
    booktitle = {Robotics: Science and Systems VIII},
    publisher = {The MIT Press},
    year = {2013},
    month = {07},
    doi = {10.7551/mitpress/9816.003.0015},
}

@misc{zhu2021challengesoutlookroboticmanipulation,
      title={Challenges and Outlook in Robotic Manipulation of Deformable Objects}, 
      author={Jihong Zhu and Andrea Cherubini and Claire Dune and others},
      year={2021},
      eprint={2105.01767},
      archivePrefix={arXiv},
      primaryClass={cs.RO}, 
}

@inproceedings{grc,
  author ="Guler, B. and  Pompetzki, K. and  Manschitz, S. and  Peters, J.",
  year =		 "2025",
  title =		 "Towards Assistive Teleoperation for Knot Untangling",
  booktitle =		 "German Robotics Conference (GRC)",
  key =			 "teleoperation, shared autonomy, deformable linear object",
}

@incollection{hart1988development,
title = {Development of NASA-TLX (Task Load Index): Results of Empirical and Theoretical Research},
editor = {Peter A. Hancock and Najmedin Meshkati},
series = {Advances in Psychology},
publisher = {North-Holland},
volume = {52},
pages = {139-183},
year = {1988},
booktitle = {Human Mental Workload},
issn = {0166-4115},
doi = {https://doi.org/10.1016/S0166-4115(08)62386-9},
author = {Sandra G. Hart and Lowell E. Staveland}
}

@INPROCEEDINGS{shivakumarSGTM20Autonomously2022,
  author={Shivakumar, Kaushik and Viswanath, Vainavi and Gu, Anrui and others},
  booktitle={2023 IEEE International Conference on Robotics and Automation (ICRA)}, 
  title={SGTM 2.0: Autonomously Untangling Long Cables using Interactive Perception}, 
  year={2023},
  volume={},
  number={},
  pages={5837-5843},
  doi={10.1109/ICRA48891.2023.10160574}}

@article{selvaggioAutonomyPhysicalHumanRobot2021,
	title = {Autonomy in {Physical} {Human}-{Robot} {Interaction}: {A} {Brief} {Survey}},
	volume = {6},
	issn = {2377-3766, 2377-3774},
	shorttitle = {Autonomy in {Physical} {Human}-{Robot} {Interaction}},
	doi = {10.1109/LRA.2021.3100603},
	language = {en},
	number = {4},
	urldate = {2024-02-01},
	journal = {IEEE Robotics and Automation Letters},
	author = {Selvaggio, Mario and Cognetti, Marco and Nikolaidis, Stefanos and Ivaldi, Serena and Siciliano, Bruno},
	month = oct,
	year = {2021},
	pages = {7989--7996},
}

@article{muellingAutonomyInfusedTeleoperation2015,
author = {Muelling, Katharina and Venkatraman, Arun and Valois, Jean-Sebastien and others},
title = {Autonomy infused teleoperation with application to brain computer interface controlled manipulation},
year = {2017},
issue_date = {August    2017},
publisher = {Kluwer Academic Publishers},
address = {USA},
volume = {41},
number = {6},
issn = {0929-5593},
doi = {10.1007/s10514-017-9622-4},
journal = {Auton. Robots},
month = aug,
pages = {1401–1422},
numpages = {22},

}

@article{javdaniSharedAutonomyHindsight2018,
	title = {Shared autonomy via hindsight optimization for teleoperation and teaming},
	volume = {37},
	issn = {0278-3649, 1741-3176},
	doi = {10.1177/0278364918776060},
	language = {en},
	number = {7},
	urldate = {2024-02-01},
	journal = {The International Journal of Robotics Research},
	author = {Javdani, Shervin and Admoni, Henny and Pellegrinelli, Stefania and Srinivasa, Siddhartha S. and Bagnell, J. Andrew},
	month = jun,
	year = {2018},
	pages = {717--742},
}

@misc{yonedaNoiseBackDiffusion2023,
      title={To the Noise and Back: Diffusion for Shared Autonomy}, 
      author={Takuma Yoneda and Luzhe Sun and Ge Yang and Bradly Stadie and Matthew Walter},
      year={2025},
      eprint={2302.12244},
      archivePrefix={arXiv},
      primaryClass={cs.RO},
}

@InProceedings{viswanathHANDLOOMLearnedTracing2023a,
  title = 	 {HANDLOOM: Learned Tracing of One-Dimensional Objects for Inspection and Manipulation},
  author =       {Viswanath, Vainavi and Shivakumar, Kaushik and Parulekar, Mallika and others},
  booktitle = 	 {Proceedings of The 7th Conference on Robot Learning},
  pages = 	 {341--357},
  year = 	 {2023},
  editor = 	 {Tan, Jie and Toussaint, Marc and Darvish, Kourosh},
  volume = 	 {229},
  series = 	 {Proceedings of Machine Learning Research},
  month = 	 {06--09 Nov},
  publisher =    {PMLR},
}

@misc{reddySharedAutonomyDeep2018,
      title={Shared Autonomy via Deep Reinforcement Learning}, 
      author={Siddharth Reddy and Anca D. Dragan and Sergey Levine},
      year={2018},
      eprint={1802.01744},
      archivePrefix={arXiv},
      primaryClass={cs.LG},
}

@inproceedings{chiaravalliVisionbasedSharedAutonomy2023,
	address = {Seattle, WA, USA},
	title = {A {Vision}-based {Shared} {Autonomy} {Framework} for {Deformable} {Linear} {Objects} {Manipulation}},
	copyright = {https://doi.org/10.15223/policy-029},
	isbn = {978-1-6654-7633-1},
	doi = {10.1109/AIM46323.2023.10196145},
	language = {en},
	urldate = {2024-06-26},
	booktitle = {2023 {IEEE}/{ASME} {International} {Conference} on {Advanced} {Intelligent} {Mechatronics} ({AIM})},
	publisher = {IEEE},
	author = {Chiaravalli, Davide and Caporali, Alessio and Friz, Anna and Meattini, Roberto and Palli, Gianluca},
	month = jun,
	year = {2023},
	pages = {733--738},
}

@article{caporaliDeformableLinearObjects2024,
	title = {Deformable {Linear} {Objects} {Manipulation} {With} {Online} {Model} {Parameters} {Estimation}},
	volume = {9},
	copyright = {https://creativecommons.org/licenses/by/4.0/legalcode},
	issn = {2377-3766, 2377-3774},
	doi = {10.1109/LRA.2024.3357310},
	language = {en},
	number = {3},
	urldate = {2024-06-26},
	journal = {IEEE Robotics and Automation Letters},
	author = {Caporali, Alessio and Kicki, Piotr and Galassi, Kevin and Zanella, Riccardo and Walas, Krzysztof and Palli, Gianluca},
	month = mar,
	year = {2024},
	pages = {2598--2605},
}

@article{draganPolicyblendingFormalismShared2013,
	title = {A policy-blending formalism for shared control},
	volume = {32},
	issn = {0278-3649, 1741-3176},
	doi = {10.1177/0278364913490324},
	language = {en},
	number = {7},
	urldate = {2025-02-17},
	journal = {The International Journal of Robotics Research},
	author = {Dragan, Anca D and Srinivasa, Siddhartha S},
	month = jun,
	year = {2013},
	pages = {790--805},
}

@article{gopinathHumanintheLoopOptimizationShared2017,
	title = {Human-in-the-{Loop} {Optimization} of {Shared} {Autonomy} in {Assistive} {Robotics}},
	volume = {2},
	copyright = {https://ieeexplore.ieee.org/Xplorehelp/downloads/license-information/IEEE.html},
	issn = {2377-3766, 2377-3774},
	doi = {10.1109/LRA.2016.2593928},
	language = {en},
	number = {1},
	urldate = {2025-02-14},
	journal = {IEEE Robotics and Automation Letters},
	author = {Gopinath, Deepak and Jain, Siddarth and Argall, Brenna D.},
	month = jan,
	year = {2017},
	pages = {247--254},
}

@misc{belsareZeroShotUserIntent2025,
	title = {Toward {Zero}-{Shot} {User} {Intent} {Recognition} in {Shared} {Autonomy}},
	doi = {10.48550/arXiv.2501.08389},
	language = {en},
	urldate = {2025-02-14},
	publisher = {arXiv},
	author = {Belsare, Atharv and Karimi, Zohre and Mattson, Connor and Brown, Daniel S.},
	month = jan,
	year = {2025},
}

@article{caporaliRoboticManipulationDeformable2025,
author={Caporali, Alessio and Palli, Gianluca},
  journal={IEEE/ASME Transactions on Mechatronics}, 
  title={Robotic Manipulation of Deformable Linear Objects via Multiview Model-Based Visual Tracking}, 
  year={2025},
  volume={30},
  number={5},
  pages={3966-3977},
  doi={10.1109/TMECH.2025.3562295}}

@misc{dinkelKnotDLOInterpretableKnot,
      title={KnotDLO: Toward Interpretable Knot Tying}, 
      author={Holly Dinkel and Raghavendra Navaratna and Jingyi Xiang and Brian Coltin and Trey Smith and Timothy Bretl},
      year={2025},
      eprint={2506.22176},
      archivePrefix={arXiv},
      primaryClass={cs.RO},
}

@inproceedings{yuHANDLOOM30Interactive,
	title = {{HANDLOOM} 3.0: {Interactive} {Bi}-{Directional} {Cable} {Tracing} {Amid} {Clutter}},
	language = {en},
	author = {Yu, Justin and Shivakumar, Nidhya and Sumedh, Veena and others},
    booktitle = {IEEE ICRA, 5th Workshop: Reflections on Representations and Manipulating Deformable Objects},
    year = {2025},
    language = {en}
}

@article{almaghoutRoboticComanipulationDeformable2024,
	title = {Robotic co-manipulation of deformable linear objects for large deformation tasks},
	volume = {175},
	issn = {09218890},
	doi = {10.1016/j.robot.2024.104652},
	language = {en},
	urldate = {2026-02-16},
	journal = {Robotics and Autonomous Systems},
	author = {Almaghout, Karam and Cherubini, Andrea and Klimchik, Alexandr},
	month = may,
	year = {2024},
	pages = {104652},
}

@inproceedings{chenContactAwareShapingMaintenance2023,
	address = {Detroit, MI, USA},
	title = {Contact-{Aware} {Shaping} and {Maintenance} of {Deformable} {Linear} {Objects} {With} {Fixtures}},
	copyright = {https://doi.org/10.15223/policy-029},
	isbn = {978-1-6654-9190-7},
	doi = {10.1109/IROS55552.2023.10341726},
	language = {en},
	urldate = {2026-02-16},
	booktitle = {2023 {IEEE}/{RSJ} {International} {Conference} on {Intelligent} {Robots} and {Systems} ({IROS})},
	publisher = {IEEE},
	author = {Chen, Kejia and Bing, Zhenshan and Wu, Fan and others},
	month = oct,
	year = {2023},
	pages = {1--8},
}

@article{rambowAutonomousManipulationDeformable2012,
  title={Autonomous manipulation of deformable objects based on teleoperated demonstrations},
  author={Matthias Rambow and Thomas Schauss and Martin Buss and Sandra Hirche},
  journal={2012 IEEE/RSJ International Conference on Intelligent Robots and Systems},
  year={2012},
  pages={2809-2814},
  url={https://api.semanticscholar.org/CorpusID:11310079}
}

@inproceedings{garcia-camachoStandardizationClothObjects2024,
	title = {Standardization of {Cloth} {Objects} and its {Relevance} in {Robotic} {Manipulation}},
	doi = {10.1109/ICRA57147.2024.10610630},
	language = {en},
	urldate = {2026-02-16},
	booktitle = {2024 {IEEE} {International} {Conference} on {Robotics} and {Automation} ({ICRA})},
	author = {Garcia-Camacho, Irene and Longhini, Alberta and Welle, Michael and Alenyà, Guillem and Kragic, Danica and Borràs, Júlia},
	month = may,
	year = {2024},
	pages = {8298--8304},
}

@article{caporaliDeformableLinearObjects2023,
	title = {Deformable {Linear} {Objects} {3D} {Shape} {Estimation} and {Tracking} {From} {Multiple} {2D} {Views}},
	volume = {8},
	copyright = {https://creativecommons.org/licenses/by/4.0/legalcode},
	issn = {2377-3766, 2377-3774},
	doi = {10.1109/LRA.2023.3273518},
	language = {en},
	number = {6},
	urldate = {2026-02-16},
	journal = {IEEE Robotics and Automation Letters},
	author = {Caporali, Alessio and Galassi, Kevin and Palli, Gianluca},
	month = jun,
	year = {2023},
	pages = {3852--3859},
}

@article{yuGlobalModelLearning2023,
	title = {Global {Model} {Learning} for {Large} {Deformation} {Control} of {Elastic} {Deformable} {Linear} {Objects}: {An} {Efficient} and {Adaptive} {Approach}},
	volume = {39},
	copyright = {https://ieeexplore.ieee.org/Xplorehelp/downloads/license-information/IEEE.html},
	issn = {1552-3098, 1941-0468},
	shorttitle = {Global {Model} {Learning} for {Large} {Deformation} {Control} of {Elastic} {Deformable} {Linear} {Objects}},
	doi = {10.1109/TRO.2022.3200546},
	language = {en},
	number = {1},
	urldate = {2026-02-16},
	journal = {IEEE Transactions on Robotics},
	author = {Yu, Mingrui and Lv, Kangchen and Zhong, Hanzhong and Song, Shiji and Li, Xiang},
	month = feb,
	year = {2023},
	pages = {417--436},
}

@article{matsunoManipulationDeformableLinear2006,
	title = {Manipulation of deformable linear objects using knot invariants to classify the object condition based on image sensor information},
	volume = {11},
	copyright = {https://ieeexplore.ieee.org/Xplorehelp/downloads/license-information/IEEE.html},
	issn = {1083-4435},
	doi = {10.1109/TMECH.2006.878557},
	language = {en},
	number = {4},
	urldate = {2026-02-16},
	journal = {IEEE/ASME Transactions on Mechatronics},
	author = {Matsuno, T. and Tamaki, D. and Arai, F. and Fukuda, T.},
	month = aug,
	year = {2006},
	pages = {401--408},
}

@article{aksoyPlanningControlDeformable2026,
	title = {Planning and {Control} for {Deformable} {Linear} {Object} {Manipulation}},
	volume = {23},
	copyright = {https://ieeexplore.ieee.org/Xplorehelp/downloads/license-information/IEEE.html},
	issn = {1545-5955, 1558-3783},
	doi = {10.1109/TASE.2025.3635685},
	language = {en},
	urldate = {2026-02-16},
	journal = {IEEE Transactions on Automation Science and Engineering},
	author = {Aksoy, Burak and Wen, John T.},
	year = {2026},
	pages = {1093--1111},
}

@misc{raviSAM2Segment2024,
	title = {{SAM} 2: {Segment} {Anything} in {Images} and {Videos}},
	shorttitle = {{SAM} 2},
	doi = {10.48550/arXiv.2408.00714},
	language = {en},
	urldate = {2026-02-16},
	publisher = {arXiv},
	author = {Ravi, Nikhila and Gabeur, Valentin and Hu, Yuan-Ting and others},
	month = oct,
	year = {2024},
}

@article{tangLearningBasedMPCSafety2024,
	title = {Learning-{Based} {MPC} {With} {Safety} {Filter} for {Constrained} {Deformable} {Linear} {Object} {Manipulation}},
	volume = {9},
	copyright = {https://ieeexplore.ieee.org/Xplorehelp/downloads/license-information/IEEE.html},
	issn = {2377-3766, 2377-3774},
	doi = {10.1109/LRA.2024.3362643},
	language = {en},
	number = {3},
	urldate = {2026-02-16},
	journal = {IEEE Robotics and Automation Letters},
	author = {Tang, Yunxi and Chu, Xiangyu and Huang, Jing and Samuel Au, K. W.},
	month = mar,
	year = {2024},
	pages = {2877--2884},
}

@misc{aksoyCollaborativeManipulationDeformable2024,
	title = {Collaborative {Manipulation} of {Deformable} {Objects} with {Predictive} {Obstacle} {Avoidance}},
	doi = {10.48550/arXiv.2401.16560},
	language = {en},
	urldate = {2026-02-16},
	publisher = {arXiv},
	author = {Aksoy, Burak and Wen, John},
	month = jan,
	year = {2024},
}

@inproceedings{quartaroModelingDeformableLinear2024,
	address = {Miami, Florida},
	title = {Modeling {Deformable} {Linear} {Objects} for {Autonomous} {Robotic} {Outfitting} of {Lunar} {Surface} {Systems}},
	isbn = {978-0-7844-8573-6},
	doi = {10.1061/9780784485736.097},
	language = {en},
	urldate = {2026-02-16},
	booktitle = {Earth and {Space} 2024},
	publisher = {American Society of Civil Engineers},
	author = {Quartaro, Amy M. and Cooper, John R. and Moser, Joshua N. and Komendera, Erik E.},
	month = oct,
	year = {2024},
	pages = {1112--1124},
}

@misc{chenMultiRobotAssemblyDeformable2025,
	title = {Multi-{Robot} {Assembly} of {Deformable} {Linear} {Objects} {Using} {Multi}-{Modal} {Perception}},
	doi = {10.48550/arXiv.2506.22034},
	language = {en},
	urldate = {2026-02-16},
	publisher = {arXiv},
	author = {Chen, Kejia and Dettmering, Celina and Pachler, Florian and others},
	month = jun,
	year = {2025},
}

@misc{zhangParticleGridNeuralDynamics2025,
	title = {Particle-{Grid} {Neural} {Dynamics} for {Learning} {Deformable} {Object} {Models} from {RGB}-{D} {Videos}},
	doi = {10.48550/arXiv.2506.15680},
	language = {en},
	urldate = {2026-02-16},
	publisher = {arXiv},
	author = {Zhang, Kaifeng and Li, Baoyu and Hauser, Kris and Li, Yunzhu},
	month = nov,
	year = {2025},
}

@misc{grannenUntanglingDenseKnots2020,
	title = {Untangling {Dense} {Knots} by {Learning} {Task}-{Relevant} {Keypoints}},
	doi = {10.48550/arXiv.2011.04999},
	language = {en},
	urldate = {2026-02-16},
	publisher = {arXiv},
	author = {Grannen, Jennifer and Sundaresan, Priya and Thananjeyan, Brijen and others},
	month = nov,
	year = {2020},
}

@misc{sundaresanUntanglingDenseNonPlanar2021,
	title = {Untangling {Dense} {Non}-{Planar} {Knots} by {Learning} {Manipulation} {Features} and {Recovery} {Policies}},
	doi = {10.48550/arXiv.2107.08942},
	language = {en},
	urldate = {2026-02-16},
	publisher = {arXiv},
	author = {Sundaresan, Priya and Grannen, Jennifer and Thananjeyan, Brijen and others},
	month = jun,
	year = {2021},
}

@article{caporaliRTDLORealTimeDeformable2023,
	title = {{RT}-{DLO}: {Real}-{Time} {Deformable} {Linear} {Objects} {Instance} {Segmentation}},
	volume = {19},
	copyright = {https://creativecommons.org/licenses/by/4.0/legalcode},
	issn = {1551-3203, 1941-0050},
	shorttitle = {{RT}-{DLO}},
	doi = {10.1109/TII.2023.3245641},
	language = {en},
	number = {11},
	urldate = {2026-02-16},
	journal = {IEEE Transactions on Industrial Informatics},
	author = {Caporali, Alessio and Galassi, Kevin and Žagar, Bare Luka and Zanella, Riccardo and Palli, Gianluca and Knoll, Alois C},
	month = nov,
	year = {2023},
	pages = {11333--11342},
}

@misc{kozlovskyISCUTEInstanceSegmentation2024,
	title = {{ISCUTE}: {Instance} {Segmentation} of {Cables} {Using} {Text} {Embedding}},
	shorttitle = {{ISCUTE}},
	doi = {10.48550/arXiv.2402.11996},
	language = {en},
	urldate = {2026-02-16},
	publisher = {arXiv},
	author = {Kozlovsky, Shir and Joglekar, Omkar and Castro, Dotan Di},
	month = feb,
	year = {2024},
}

@misc{liGRRLGoingDexterous2025,
	title = {{GR}-{RL}: {Going} {Dexterous} and {Precise} for {Long}-{Horizon} {Robotic} {Manipulation}},
	shorttitle = {{GR}-{RL}},
	doi = {10.48550/arXiv.2512.01801},
	language = {en},
	urldate = {2026-02-16},
	publisher = {arXiv},
	author = {Li, Yunfei and Ma, Xiao and Xu, Jiafeng and others},
	month = dec,
	year = {2025},
}

@article{xiangTrackDLOTrackingDeformable2023,
	title = {{TrackDLO}: {Tracking} {Deformable} {Linear} {Objects} {Under} {Occlusion} {With} {Motion} {Coherence}},
	volume = {8},
	copyright = {https://ieeexplore.ieee.org/Xplorehelp/downloads/license-information/IEEE.html},
	issn = {2377-3766, 2377-3774},
	shorttitle = {{TrackDLO}},
	doi = {10.1109/LRA.2023.3303710},
	language = {en},
	number = {10},
	urldate = {2026-02-16},
	journal = {IEEE Robotics and Automation Letters},
	author = {Xiang, Jingyi and Dinkel, Holly and Zhao, Harry and others},
	month = oct,
	year = {2023},
	pages = {6179--6186},
}

@misc{yanSelfSupervisedLearningState2020,
	title = {Self-{Supervised} {Learning} of {State} {Estimation} for {Manipulating} {Deformable} {Linear} {Objects}},
	doi = {10.48550/arXiv.1911.06283},
	language = {en},
	urldate = {2026-02-16},
	publisher = {arXiv},
	author = {Yan, Mengyuan and Zhu, Yilin and Jin, Ning and Bohg, Jeannette},
	month = oct,
	year = {2020},
}

@misc{zhangHarnessingTwistingSingleArm2024,
	title = {Harnessing with {Twisting}: {Single}-{Arm} {Deformable} {Linear} {Object} {Manipulation} for {Industrial} {Harnessing} {Task}},
	shorttitle = {Harnessing with {Twisting}},
	doi = {10.48550/arXiv.2410.10729},
	language = {en},
	urldate = {2026-02-16},
	publisher = {arXiv},
	author = {Zhang, Xiang and Lin, Hsien-Chung and Zhao, Yu and Tomizuka, Masayoshi},
	month = oct,
	year = {2024},
}

@article{zhaoleRobustDeformableLinear2024,
	title = {A {Robust} {Deformable} {Linear} {Object} {Perception} {Pipeline} in {3D}: {From} {Segmentation} to {Reconstruction}},
	volume = {9},
	copyright = {https://ieeexplore.ieee.org/Xplorehelp/downloads/license-information/IEEE.html},
	issn = {2377-3766, 2377-3774},
	shorttitle = {A {Robust} {Deformable} {Linear} {Object} {Perception} {Pipeline} in {3D}},
	doi = {10.1109/LRA.2023.3337695},
	language = {en},
	number = {1},
	urldate = {2026-02-18},
	journal = {IEEE Robotics and Automation Letters},
	author = {Zhaole, Sun and Zhou, Hang and Nanbo, Li and Chen, Longfei and Zhu, Jihong and Fisher, Robert B.},
	month = jan,
	year = {2024},
	pages = {843--850},
}

@article{huangLearningGraphDynamics2023,
	title = {Learning {Graph} {Dynamics} {With} {External} {Contact} for {Deformable} {Linear} {Objects} {Shape} {Control}},
	volume = {8},
	copyright = {https://ieeexplore.ieee.org/Xplorehelp/downloads/license-information/IEEE.html},
	issn = {2377-3766, 2377-3774},
	doi = {10.1109/LRA.2023.3264764},
	language = {en},
	number = {6},
	urldate = {2026-02-18},
	journal = {IEEE Robotics and Automation Letters},
	author = {Huang, Yichen and Xia, Chongkun and Wang, Xueqian and Liang, Bin},
	month = jun,
	year = {2023},
	pages = {3892--3899},
}

@inproceedings{benschPhysicsInformedNeuralNetworks2024,
	address = {Yokohama, Japan},
	title = {Physics-{Informed} {Neural} {Networks} for {Continuum} {Robots}: {Towards} {Fast} {Approximation} of {Static} {Cosserat} {Rod} {Theory}},
	copyright = {https://doi.org/10.15223/policy-029},
	isbn = {979-8-3503-8457-4},
	shorttitle = {Physics-{Informed} {Neural} {Networks} for {Continuum} {Robots}},
	doi = {10.1109/ICRA57147.2024.10610742},
	language = {en},
	urldate = {2026-02-18},
	booktitle = {2024 {IEEE} {International} {Conference} on {Robotics} and {Automation} ({ICRA})},
	publisher = {IEEE},
	author = {Bensch, Martin and Job, Tim-David and Habich, Tim-Lukas and Seel, Thomas and Schappler, Moritz},
	month = may,
	year = {2024},
	pages = {17293--17299},
}

@article{sanchezRoboticManipulationSensing2018,
	title = {Robotic manipulation and sensing of deformable objects in domestic and industrial applications: a survey},
	volume = {37},
	issn = {0278-3649, 1741-3176},
	shorttitle = {Robotic manipulation and sensing of deformable objects in domestic and industrial applications},
	doi = {10.1177/0278364918779698},
	language = {en},
	number = {7},
	urldate = {2026-02-18},
	journal = {The International Journal of Robotics Research},
	author = {Sanchez, Jose and Corrales, Juan-Antonio and Bouzgarrou, Belhassen-Chedli and Mezouar, Youcef},
	month = jun,
	year = {2018},
	pages = {688--716},
}

@article{yinModelingLearningPerception2021,
	title = {Modeling, learning, perception, and control methods for deformable object manipulation},
	volume = {6},
	issn = {2470-9476},
	doi = {10.1126/scirobotics.abd8803},
	language = {en},
	number = {54},
	urldate = {2026-02-18},
	journal = {Science Robotics},
	author = {Yin, Hang and Varava, Anastasia and Kragic, Danica},
	month = may,
	year = {2021},
	pages = {eabd8803},
}

@article{wangOfflineOnlineLearningDeformation2022,
	title = {Offline-{Online} {Learning} of {Deformation} {Model} for {Cable} {Manipulation} with {Graph} {Neural} {Networks}},
	volume = {7},
	issn = {2377-3766, 2377-3774},
	doi = {10.1109/LRA.2022.3158376},
	number = {2},
	urldate = {2026-02-18},
	journal = {IEEE Robotics and Automation Letters},
	author = {Wang, Changhao and Zhang, Yuyou and Zhang, Xiang and Wu, Zheng and Zhu, Xinghao and Jin, Shiyu and Tang, Te and Tomizuka, Masayoshi},
	month = apr,
	year = {2022},
	pages = {5544--5551},
}

@inproceedings{leeLearningForcebasedManipulation2015,
	address = {Seattle, WA, USA},
	title = {Learning force-based manipulation of deformable objects from multiple demonstrations},
	isbn = {978-1-4799-6923-4},
	doi = {10.1109/ICRA.2015.7138997},
	urldate = {2026-02-18},
	booktitle = {2015 {IEEE} {International} {Conference} on {Robotics} and {Automation} ({ICRA})},
	publisher = {IEEE},
	author = {Lee, Alex X. and Lu, Henry and Gupta, Abhishek and Levine, Sergey and Abbeel, Pieter},
	month = may,
	year = {2015},
	pages = {177--184},
}

@inproceedings{adebolaAutomatingDeformableGasket2024,
	address = {Bari, Italy},
	title = {Automating {Deformable} {Gasket} {Assembly}},
	copyright = {https://doi.org/10.15223/policy-029},
	isbn = {979-8-3503-5851-3},
	doi = {10.1109/CASE59546.2024.10711836},
	urldate = {2026-02-18},
	booktitle = {2024 {IEEE} 20th {International} {Conference} on {Automation} {Science} and {Engineering} ({CASE})},
	publisher = {IEEE},
	author = {Adebola, Simeon and Sadjadpour, Tara and El-Refai, Karim and others},
	month = aug,
	year = {2024},
	pages = {4146--4153},
}

@inproceedings{kickiDLOFTBsFastTracking2023,
  author={Kicki, Piotr and Szymko, Amadeusz and Walas, Krzysztof},
  booktitle={2023 IEEE International Conference on Robotics and Automation (ICRA)}, 
  title={DLOFTBs – Fast Tracking of Deformable Linear Objects with B-splines}, 
  year={2023},
  volume={},
  number={},
  pages={7104-7110},
  doi={10.1109/ICRA48891.2023.10160437}}

@article{luoTSLTrackingDeformable2025,
	title = {{TSL}: {Tracking} {Deformable} {Linear} {Objects} for {Bimanual} {Shoe} {Lacing}},
	volume = {10},
	copyright = {https://ieeexplore.ieee.org/Xplorehelp/downloads/license-information/IEEE.html},
	issn = {2377-3766, 2377-3774},
	shorttitle = {{TSL}},
	doi = {10.1109/LRA.2025.3583476},
	number = {8},
	urldate = {2026-02-18},
	journal = {IEEE Robotics and Automation Letters},
	author = {Luo, Haining and Demiris, Yiannis},
	month = aug,
	year = {2025},
	pages = {8212--8219},
}

@article{kitagawaEffectSensorySubstitution2005,
  title = {Effect of Sensory Substitution on Suture-Manipulation Forces for Robotic Surgical Systems},
  author = {Kitagawa, Masaya and Dokko, Daniell and Okamura, Allison M. and Yuh, David D.},
  year = 2005,
  month = jan,
  journal = {The Journal of Thoracic and Cardiovascular Surgery},
  volume = {129},
  number = {1},
  pages = {151--158},
  issn = {00225223},
  doi = {10.1016/j.jtcvs.2004.05.029},
  urldate = {2026-02-19},
  copyright = {https://www.elsevier.com/tdm/userlicense/1.0/},
  langid = {english},
}

@article{shahAircraftAssembly2018,
  title = {Planning for Manipulation of Interlinked Deformable Linear Objects With Applications to Aircraft Assembly},
  author = {Shah, Ankit and Blumberg, Lotta and Shah, Julie},
  year = 2018,
  month = oct,
  journal = {IEEE Transactions on Automation Science and Engineering},
  volume = {15},
  number = {4},
  pages = {1823--1838},
  issn = {1545-5955, 1558-3783},
  doi = {10.1109/TASE.2018.2811626},
  urldate = {2026-02-20},
  copyright = {https://ieeexplore.ieee.org/Xplorehelp/downloads/license-information/IEEE.html},
  langid = {english},
}

@article{salunkhe2023specifying,
title = {Specifying task allocation in automotive wire harness assembly stations for Human-Robot Collaboration},
author = {Omkar Salunkhe and Johan Stahre and David Romero and Dan Li and Björn Johansson},
journal = {Computers \& Industrial Engineering},
volume = {184},
pages = {109572},
year = {2023},
issn = {0360-8352},
doi = {https://doi.org/10.1016/j.cie.2023.109572},
}

@inproceedings{arevaloarboledaAssistingManipulationGrasping2021,
  title = {Assisting {{Manipulation}} and {{Grasping}} in {{Robot Teleoperation}} with {{Augmented Reality Visual Cues}}},
  booktitle = {Proceedings of the 2021 {{CHI Conference}} on {{Human Factors}} in {{Computing Systems}}},
  author = {Arevalo Arboleda, Stephanie and R{\"u}cker, Franziska and Dierks, Tim and Gerken, Jens},
  year = 2021,
  month = may,
  pages = {1--14},
  publisher = {ACM},
  address = {Yokohama Japan},
  doi = {10.1145/3411764.3445398},
  urldate = {2026-02-22},
  isbn = {978-1-4503-8096-6},
  langid = {english},
}

@inproceedings{krishnanHumanPreferredAugmented2023,
  title = {Human {{Preferred Augmented Reality Visual Cues}} for {{Remote Robot Manipulation Assistance}}: From {{Direct}} to {{Supervisory Control}}},
  shorttitle = {Human {{Preferred Augmented Reality Visual Cues}} for {{Remote Robot Manipulation Assistance}}},
  booktitle = {2023 {{IEEE}}/{{RSJ International Conference}} on {{Intelligent Robots}} and {{Systems}} ({{IROS}})},
  author = {Krishnan, Achyuthan Unni and Lin, Tsung-Chi and Li, Zhi},
  year = 2023,
  month = oct,
  pages = {7034--7039},
  publisher = {IEEE},
  address = {Detroit, MI, USA},
  doi = {10.1109/IROS55552.2023.10341969},
  urldate = {2026-02-22},
  copyright = {https://doi.org/10.15223/policy-029},
  isbn = {978-1-6654-9190-7},
  langid = {english},
}

@article{linPerceptionActionAugmentation2024,
  title = {Perception and {{Action Augmentation}} for {{Teleoperation Assistance}} in {{Freeform Telemanipulation}}},
  author = {Lin, Tsung-Chi and Krishnan, Achyuthan Unni and Li, Zhi},
  year = 2024,
  month = mar,
  journal = {ACM Transactions on Human-Robot Interaction},
  volume = {13},
  number = {1},
  pages = {1--40},
  issn = {2573-9522},
  doi = {10.1145/3643804},
  urldate = {2026-02-22},
  langid = {english},
}

@article{barba2022remote,
  title={Remote telesurgery in humans: a systematic review},
  author={Barba, Patrick and Stramiello, Joshua and Funk, Emily K and Richter, Florian and Yip, Michael C and Orosco, Ryan K},
  journal={Surgical endoscopy},
  volume={36},
  number={5},
  pages={2771--2777},
  year={2022},
  publisher={Springer},
  doi = {10.1007/s00464-022-09074-4}
}

@article{muellerPBD2007,
  title        = {Position Based Dynamics},
  author       = {M{\"u}ller, Matthias and Heidelberger, Bruno and Hennix, Marcus and Ratcliff, John},
  journal      = {Journal of Visual Communication and Image Representation},
  volume       = {18},
  number       = {2},
  pages        = {109--118},
  year         = {2007},
  doi          = {10.1016/j.jvcir.2007.01.005}
  }

@misc{longSelfsupervisedPhysicsInformedManipulation2026,
  title = {Self-Supervised {{Physics-Informed Manipulation}} of {{Deformable Linear Objects}} with {{Non-negligible Dynamics}}},
  author = {Long, Youyuan and Solak, Gokhan and Zeynalpour, Sara and Zhang, Heng and Ajoudani, Arash},
  year = 2026,
  month = feb,
  number = {arXiv:2602.03623},
  eprint = {2602.03623},
  primaryclass = {cs},
  publisher = {arXiv},
  doi = {10.48550/arXiv.2602.03623},
  urldate = {2026-02-24},
  archiveprefix = {arXiv},
}

@inproceedings{provotDeformationConstraintsMass,
 author = {Xavier Provot},
 title = {Deformation Constraints in a Mass-Spring Model to Describe Rigid Cloth Behaviour},
 booktitle = {Proceedings of Graphics Interface '95},
 series = {GI '95},
 year = {1995},
 isbn = {0-9695338-4-5},
 issn = {0713-5424},
 location = {Quebec, Quebec, Canada},
 pages = {147--154},
 numpages = {8},
 publisher = {Canadian Human-Computer Communications Society},
 address = {Toronto, Ontario, Canada},
}

@misc{berk2026barrierik,
      title={A Safety-Aware Shared Autonomy Framework with BarrierIK Using Control Barrier Functions}, 
      author={Berk Guler and Kay Pompetzki and Yuanzheng Sun and Simon Manschitz and Jan Peters},
      year={2026},
      eprint={2603.01705},
      archivePrefix={arXiv},
      primaryClass={cs.RO},

}

@ARTICLE{uzunArbitrationControlBarrier2025,
  author={Uzun, M. Yusuf and Yildiz, Yildiray},
  journal={IEEE Control Systems Letters}, 
  title={Arbitration With Control Barrier Functions for Safe Shared Control}, 
  year={2025},
  volume={9},
  number={},
  pages={2789-2794},
  doi={10.1109/LCSYS.2025.3642534}}

@inproceedings{hebriTeleoperationFrameworkRobots2023a,
  title = {A {{Teleoperation Framework}} for {{Robots Utilizing Control Barrier Functions}} in {{Virtual Reality}}},
  booktitle = {Proceedings of the 16th {{International Conference}} on {{PErvasive Technologies Related}} to {{Assistive Environments}}},
  author = {Hebri, Aref and Acharya, Sneh and Theofanidis, Michail and Makedon, Fillia},
  year = 2023,
  month = jul,
  pages = {408--412},
  publisher = {ACM},
  address = {Corfu Greece},
  doi = {10.1145/3594806.3596522},
  urldate = {2026-02-26},
  isbn = {979-8-4007-0069-9},
  langid = {english},
}

@inproceedings{aronsonInferringGoalsGaze2021,
  title = {Inferring {{Goals}} with {{Gaze}} during {{Teleoperated Manipulation}}},
  booktitle = {2021 {{IEEE}}/{{RSJ International Conference}} on {{Intelligent Robots}} and {{Systems}} ({{IROS}})},
  author = {Aronson, Reuben M. and Almutlak, Nadia and Admoni, Henny},
  year = 2021,
  month = sep,
  pages = {7307--7314},
  publisher = {IEEE},
  address = {Prague, Czech Republic},
  doi = {10.1109/IROS51168.2021.9636551},
  urldate = {2026-02-28},
  copyright = {https://ieeexplore.ieee.org/Xplorehelp/downloads/license-information/IEEE.html},
  isbn = {978-1-6654-1714-3},
}

@misc{manschitzSamplingBasedGraspCollision2025,
  title = {Sampling-{{Based Grasp}} and {{Collision Prediction}} for {{Assisted Teleoperation}}},
  author = {Manschitz, Simon and Gueler, Berk and Ma, Wei and Ruiken, Dirk},
  year = 2025,
  month = apr,
  number = {arXiv:2504.18186},
  eprint = {2504.18186},
  primaryclass = {cs},
  publisher = {arXiv},
  doi = {10.48550/arXiv.2504.18186},
  urldate = {2026-03-01},
  archiveprefix = {arXiv},
}

@article{qinHapticSharedControl2025,
  title = {Haptic {{Shared Control Framework}} with {{Interaction Force Constraint Based}} on {{Control Barrier Function}} for {{Teleoperation}}},
  author = {Qin, Wenlei and Yi, Haoran and Fan, Zhibin and Zhao, Jie},
  year = 2025,
  month = jan,
  journal = {Sensors},
  volume = {25},
  number = {2},
  pages = {405},
  issn = {1424-8220},
  doi = {10.3390/s25020405},
  urldate = {2026-02-16},
  langid = {english},
}

@article{ames2019cbf,
  author       = {Aaron D. Ames and
                  Samuel Coogan and
                  Magnus Egerstedt and
                  Gennaro Notomista and
                  Koushil Sreenath and
                  Paulo Tabuada},
  title        = {Control Barrier Functions: Theory and Applications},
  journal      = {CoRR},
  volume       = {abs/1903.11199},
  year         = {2019},
  eprinttype    = {arXiv},
  eprint       = {1903.11199},
}

@INPROCEEDINGS{zhang2021cbfteleop,
  author={Zhang, Dawei and Tron, Roberto and Khurshid, Rebecca P.},
  booktitle={2021 IEEE International Conference on Robotics and Automation (ICRA)}, 
  title={Haptic Feedback Improves Human-Robot Agreement and User Satisfaction in Shared-Autonomy Teleoperation}, 
  year={2021},
  volume={},
  number={},
  pages={3306-3312},
  doi={10.1109/ICRA48506.2021.9560991}}

@article{livatino2021MRVA,
  title = {Intuitive {{Robot Teleoperation Through Multi-Sensor Informed Mixed Reality Visual Aids}}},
  author = {Livatino, Salvatore and Guastella, Dario C. and Muscato, Giovanni and Rinaldi, Vincenzo and Cantelli, Luciano and Melita, Carmelo D. and Caniglia, Alessandro and Mazza, Riccardo and Padula, Gianluca},
  year = 2021,
  journal = {IEEE Access},
  volume = {9},
  pages = {25795--25808},
  issn = {2169-3536},
  doi = {10.1109/ACCESS.2021.3057808},
  urldate = {2026-03-01},
  copyright = {https://creativecommons.org/licenses/by/4.0/legalcode},
  langid = {english},
}

@article{niPointCloudAugmented2018,
  title = {Point Cloud Augmented Virtual Reality Environment with Haptic Constraints for Teleoperation},
  author = {Ni, Dejing and Nee, Ayc and Ong, Sk and Li, Huijun and Zhu, Chengcheng and Song, Aiguo},
  year = 2018,
  month = nov,
  journal = {Transactions of the Institute of Measurement and Control},
  volume = {40},
  number = {15},
  pages = {4091--4104},
  issn = {0142-3312, 1477-0369},
  doi = {10.1177/0142331217739953},
  urldate = {2026-03-01},
  langid = {english},
}

@INPROCEEDINGS{Rusu_ICRA2011_PCL,
  author={Rusu, Radu Bogdan and Cousins, Steve},
  booktitle={2011 IEEE International Conference on Robotics and Automation}, 
  title={3D is here: Point Cloud Library (PCL)}, 
  year={2011},
  volume={},
  number={},
  pages={1-4},
  doi={10.1109/ICRA.2011.5980567}}

@inproceedings{schuhmann2022laionb,
  title={{LAION}-5B: An open large-scale dataset for training next generation image-text models},
  author={Christoph Schuhmann and
          Romain Beaumont and
          Richard Vencu and others},
  booktitle={Thirty-sixth Conference on Neural Information Processing Systems Datasets and Benchmarks Track},
  year={2022},
}

@article{norman2010likert,
  title={Likert scales, levels of measurement and the ``laws'' of statistics},
  author={Norman, Geoff},
  journal={Advances in health sciences education : theory and practice},
  volume={15},
  number={5},
  pages={625--632},
  year={2010},
  DOI={10.1007/s10459-010-9222-y},
  publisher={Springer}
}

@article{zhou_reactive_2024,
	title = {Reactive human–robot collaborative manipulation of deformable linear objects using a new topological latent control model},
	volume = {88},
	issn = {07365845},
	doi = {10.1016/j.rcim.2024.102727},
	language = {en},
	urldate = {2026-03-09},
	journal = {Robotics and Computer-Integrated Manufacturing},
	author = {Zhou, Peng and Zheng, Pai and Qi, Jiaming and others},
	month = aug,
	year = {2024},
	pages = {102727},
}

@article{wu_mixed_2026,
	title = {A mixed reality-assisted human-to-robot skill transfer approach for contact-rich assembly via visuomotor primitives},
	volume = {99},
	issn = {07365845},
	doi = {10.1016/j.rcim.2025.103208},
	language = {en},
	urldate = {2026-03-09},
	journal = {Robotics and Computer-Integrated Manufacturing},
	author = {Wu, Duidi and Zhao, Qianyou and Shen, Yuliang and others},
	month = jun,
	year = {2026},
	pages = {103208},
}

@article{pan_augmented_2021,
	title = {Augmented reality-based robot teleoperation system using {RGB}-{D} imaging and attitude teaching device},
	volume = {71},
	issn = {07365845},
	doi = {10.1016/j.rcim.2021.102167},
	language = {en},
	urldate = {2026-03-09},
	journal = {Robotics and Computer-Integrated Manufacturing},
	author = {Pan, Yong and Chen, Chengjun and Li, Dongnian and Zhao, Zhengxu and Hong, Jun},
	month = oct,
	year = {2021},
	pages = {102167},
}

@article{su_mixed_2022,
	title = {Mixed reality-integrated {3D}/{2D} vision mapping for intuitive teleoperation of mobile manipulator},
	volume = {77},
	issn = {07365845},
	doi = {10.1016/j.rcim.2022.102332},
	language = {en},
	urldate = {2026-03-09},
	journal = {Robotics and Computer-Integrated Manufacturing},
	author = {Su, Yunpeng and Chen, Xiaoqi and Zhou, Tony and Pretty, Christopher and Chase, Geoffrey},
	month = oct,
	year = {2022},
	pages = {102332},
}

@ARTICLE{otsu,
  author={Otsu, Nobuyuki},
  journal={IEEE Transactions on Systems, Man, and Cybernetics}, 
  title={A Threshold Selection Method from Gray-Level Histograms}, 
  year={1979},
  volume={9},
  number={1},
  pages={62-66},
  doi={10.1109/TSMC.1979.4310076}}

@article{Bickley01031934,
author = {W.G. Bickley},
title = {L. The heavy elastica},
journal = {The London, Edinburgh, and Dublin Philosophical Magazine and Journal of Science},
volume = {17},
number = {113},
pages = {603--622},
year = {1934},
publisher = {Taylor \& Francis},
doi = {10.1080/14786443409462419},
}
\bibliographystyle{ieeetr}

\end{document}